\documentclass[journal,transmag,twocolumn]{IEEEtran}

\usepackage{times}
\usepackage{epsfig}
\usepackage{graphicx}
\usepackage{amsmath}
\usepackage{amssymb}
\usepackage[pagebackref=true,breaklinks=true,colorlinks,bookmarks=false]{hyperref} 

\usepackage{booktabs}       

\usepackage{graphicx} 
\usepackage{amsfonts} 
\usepackage{amsmath,soul}
\usepackage{amssymb}
\usepackage[capitalize]{cleveref} 
\newcommand{\unit}[1]{\ensuremath{\, \mathrm{#1}}} 

\usepackage{dsfont} 
\usepackage[font=small,skip=0pt]{subcaption} 
\usepackage{balance}
\usepackage{bm} 
\usepackage{wrapfig}
\usepackage{float} 

\usepackage{makecell}

\usepackage{color}
\usepackage[svgnames]{xcolor} \definecolor{DarkGreen}{rgb}{0,0.5,0}
\definecolor{DarkRed}{rgb}{0.75,0,0}

\newcommand{\cmmnt}[1]{\ignorespaces}

\newcommand{\bit}{\begin{itemize}}
\newcommand{\ei}{\end{itemize}}

\makeatletter
\renewcommand\paragraph{\@startsection{subsubsection}{4}{\z@}%
{0.25ex \@plus.5ex \@minus.2ex}%
{-.15em}%
{\normalfont\normalsize\itshape}}
\makeatother

 \usepackage[normalem]{ulem} 

\usepackage[authormarkuptext=name,addedmarkup=bf,authormarkupposition=left]{changes}
\definechangesauthor[name={X.~X.}, color={orange}]{xx}
\definechangesauthor[name={Y.~Y.}, color={blue}]{yy}
\definechangesauthor[name={SRC}, color={brown}]{salva}
\definechangesauthor[name={B.~L.}, color={MediumSeaGreen}]{bl}
\begin{document}

\title{Generating Physically-Consistent Satellite Imagery \\for Climate Visualizations}

\author{\IEEEauthorblockN{Bj{\"o}rn L{\"u}tjens, 
Brandon Leshchinskiy, 
Oc\'eane Boulais$^{*}$, 
Farrukh Chishtie$^{*}$, 
\\
Natalia D\'iaz-Rodr\'iguez$^{*}$, 
Margaux Masson-Forsythe$^{*}$, 
Ana Mata-Payerro$^{*}$, 
Christian Requena-Mesa$^{*}$, 
\\
Aruna Sankaranarayanan$^{*}$, 
Aaron Pi{\~n}a, 
Yarin Gal, 
Chedy Ra\"issi, 
Alexander Lavin, 
Dava Newman
}
\\

\thanks{\hrulefill

Corresponding author: B. L{\"u}tjens; \texttt{lutjens@mit.edu}. *indicates equal contribution. } 
\thanks{
BL{\"u}, BLe, AMP and DN are with the Massachusetts Institute of Technology,
CRM with Max Planck Institute for Biogeochemistry,
FC with The University of British Columbia and Peaceful Society, Science and Innovation Foundation,
NDR with University of Granada,
OB with Scripps Institute of Oceanography.
MMF with Surgical Data Science Collective,
AS with MIT CSAIL,
AP with US Forest Service,
YG with Oxford University,
CR with Riot Games, and
AL with Pasteur Labs.}
}

\markboth{IEEE TRANSACTIONS ON GEOSCIENCE AND REMOTE SENSING (IN PRINT)}%
{Lütjens \MakeLowercase{\textit{et al.}}: Generating Physically-Consistent Satellite Imagery for Climate Visualizations}
%

\def\eie/{\textit{Earth Intelligence Engine}}

\maketitle

\begin{abstract}
Deep generative vision models are now able to synthesize realistic-looking satellite imagery. But, the possibility of hallucinations prevents their adoption for risk-sensitive applications, such as generating materials for communicating climate change. To demonstrate this issue, we train a generative adversarial network (pix2pixHD) to create synthetic satellite imagery of future flooding and reforestation events. We find that a pure deep learning-based model can generate photorealistic flood visualizations but hallucinates floods at locations that were not susceptible to flooding. To address this issue, we propose to condition and evaluate generative vision models on segmentation maps of physics-based flood models. We show that our physics-conditioned model outperforms the pure deep learning-based model and a handcrafted baseline. We evaluate the generalization capability of our method to different remote sensing data and different climate-related events (reforestation). We publish our code and dataset which includes the data for a third case study of melting Arctic sea ice and $>$30,000 labeled HD image triplets -- or the equivalent of 5.5 million images at 128x128 pixels -- for segmentation guided image-to-image translation in Earth observation. Code and data is available at \href{https://github.com/blutjens/eie-earth-public}{github.com/blutjens/eie-earth-public}.
\end{abstract}
\begin{IEEEkeywords}
Deep Generative Vision Models, Generative AI, Physics-informed Machine Learning, Generative Adversarial Networks, Image to Image Translation, Climate Change, Flooding, Reforestation, Visualization, Synthetic Satellite Imagery, Remote Sensing.
\end{IEEEkeywords}


\IEEEdisplaynontitleabstractindextext

%
\IEEEpeerreviewmaketitle


  
\section{Introduction}\label{sec:eie_intro}
\IEEEPARstart{W}{}ith climate change, natural disasters are becoming more intense~\cite{IPCC_2018}. Floods are the most frequent weather-related disaster~\cite{Cred_2015} and already cost the U.S. $4.1\unit{B \; USD}$ per year~\cite{NoaaNcei_2023}; this damage is projected to grow over the next decades~\cite{IPCC_2018}. 

\begin{figure}[t]
  \vspace{-.2in}
  \centering
      \includegraphics [trim=0 .3 0 0, clip,  width=1.\columnwidth, angle = 0]{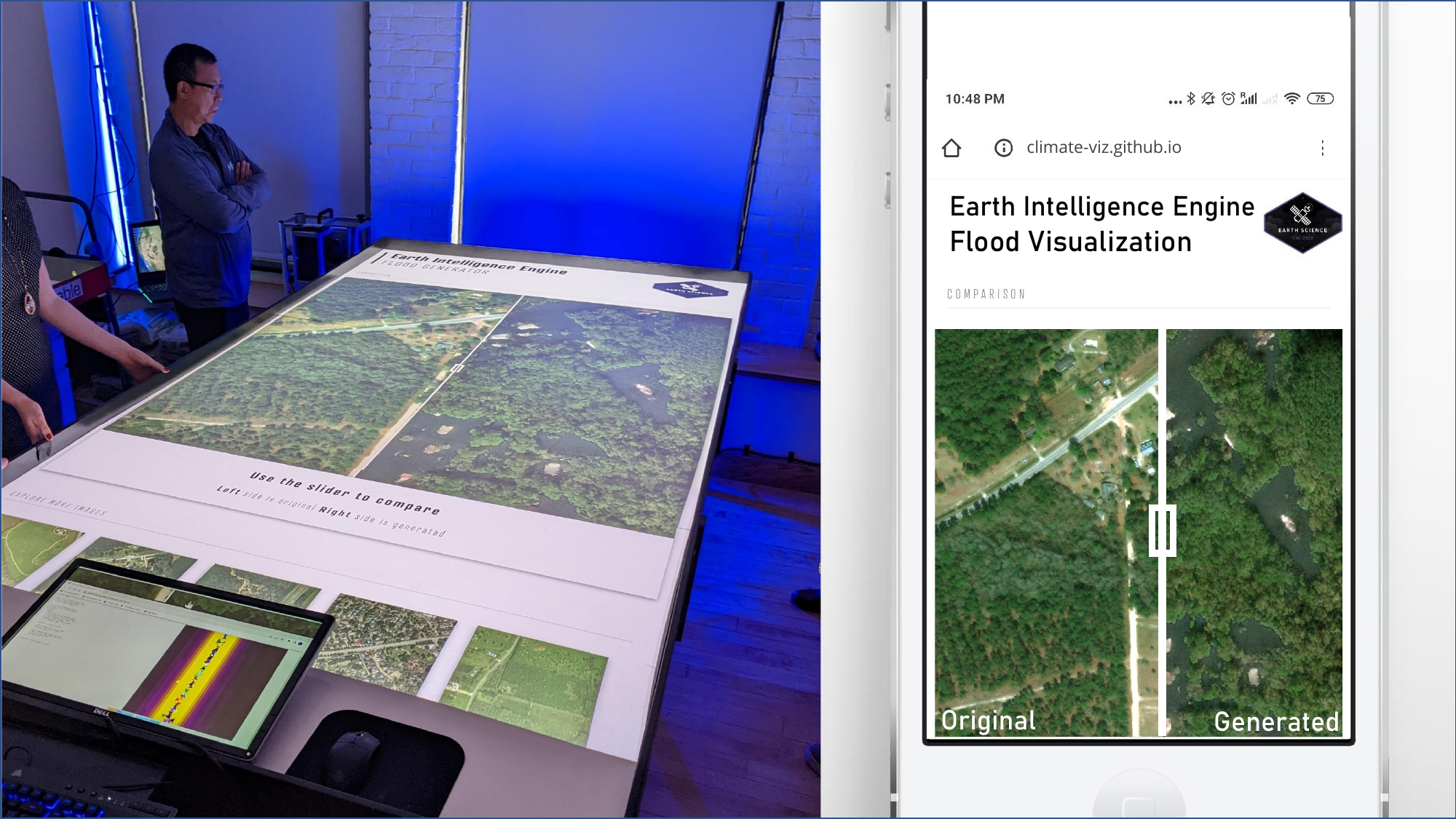}
      \vspace{-.2in}
\caption[Satellite imagery from the future]{We synthesize satellite imagery that visualizes flooding (right). We designed the underlying generative vision model to project flooding only in locations that are consistent with a physics-based flood model. The new visualizations could facilitate intuitive and trustworthy communication of climate risks, for example, via tabletop exercises as seen on the left. 
Explore more results at \href{https://climate-viz.github.io/}{climate-viz.github.io}.} \label{fig:teaserfigure} 
\end{figure}

\begin{figure}[t]
  \centering
    \subfloat[Model Architecture]{
  \includegraphics [trim=0 0 0 0, clip, width=0.50\linewidth, angle = 0]{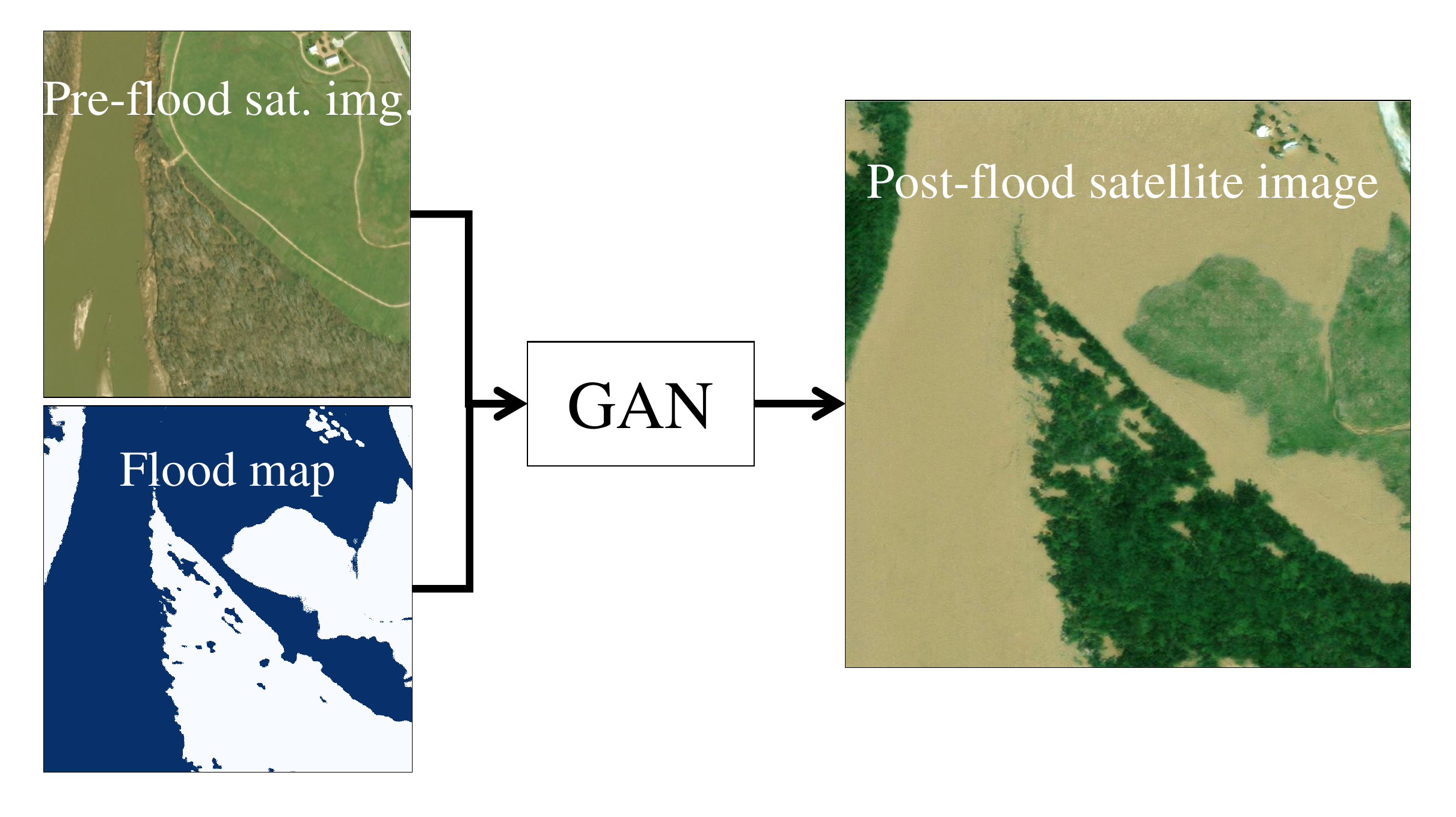}
  \label{fig:eie_model_architecture_flood}}
    \subfloat[Future extension to Arctic sea ice melt]{
    \includegraphics [trim=0 0 0 0, clip, width=0.42\linewidth, angle = 0]{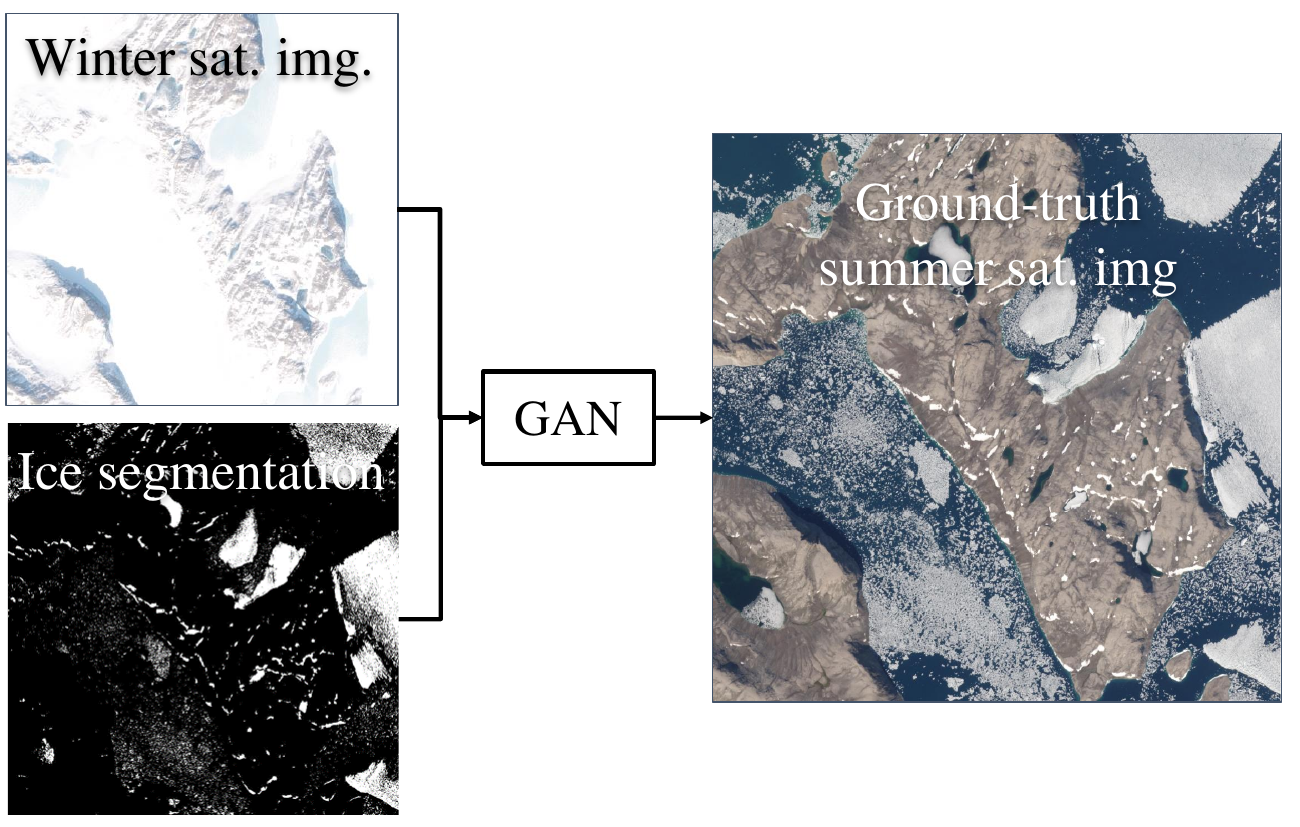}
    \label{fig:eie_model_architecture_arctic}}
\caption[Earth Intelligence Engine Model Architecture]{\textbf{Top: Model Architecture.} Our model leverages the semantic image synthesis model, Pix2pixHD~\cite{wang18pix2pixhd}, and combines a pre-flood satellite image with a physics-based flood map to generate post-flood satellite imagery. \textbf{Bottom: Arctic sea ice melt.} We publish a dataset of 19446 labeled image triplets for segmentation guided image-to-image translation which includes an additional case study on melting Arctic sea ice.} 
\label{fig:eie_model_architecture} 
\end{figure}

\begin{figure}[t]
  \vspace{-.2in}
  \centering
      \includegraphics [trim=0 .3 0 0, clip, clip, width=1.\columnwidth, angle = 0]{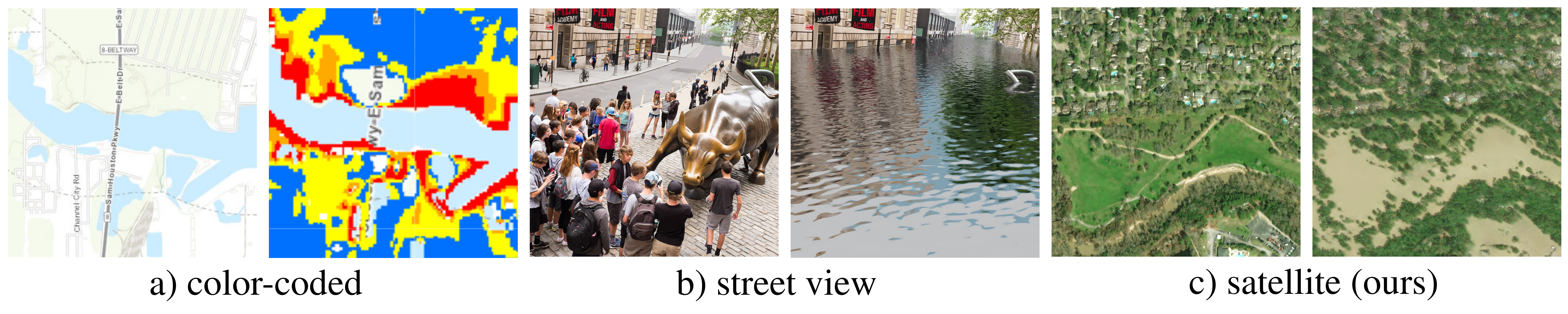}
      \vspace{-.2in}
\caption[Related flood visualizations]{\textbf{Our physically-consistent satellite imagery (c) could enable more engaging and relatable communication of city-scale flood risks~\cite{Sheppard_2012}}. Most existing visualizations of coastal floods or sea-level rise that are aimed towards the public rely on color-coded geospatial rasters (a), that can be unrelatable or impersonal~\cite{NoaaSlosh_20, FloodFactor_2020, ClimateCentral_18}. Alternative photorealistic visualizations are often limited to local street-level imagery (b)~\cite{Strauss_2015,Schmidt_2019} and lack further spatial context. Images, left-to-right:~\cite{ NoaaSlosh_20, NoaaSlosh_20, Strauss_2015, Strauss_2015, Gupta_2019}, ours.}\label{fig:related_works} 
\end{figure}

Visualizations of climate impacts are widely used by policy and decision makers to raise environmental awareness and facilitate dialogue on long-term climate adaptation decisions~\cite{Sheppard_2012}. Visualizations of flood risks, for example, are used in local policy making and community discussion groups as  decision-aids for flood infrastructure investments~\cite{Sheppard_2012}.
Current visualizations of flood impacts, however, are limited to color-coded flood maps~\cite{NoaaSlosh_20,climatecentral23viewer,noaa23sealevelviewer} or synthetic street-view imagery~\cite{Strauss_2015,Schmidt_2019}, which do not convey city-wide flood impacts in a compelling manner, as shown in~\cref{fig:related_works} and~\cite{Sheppard_2005}. 
Our work generates synthetic satellite imagery of future coastal floods that are informed by the projections of expert-validated flood models, as illustrated in~\cref{fig:eie_model_architecture}. As a geospatial rendering, this imagery might enable a more engaging communication of city-wide flood risks to governmental offices, as demoed in~\cref{fig:teaserfigure}. 


We focus on deep generative vision models, such as generative adversarial networks (GANs)~\cite{goodfellow14gans}, as our method for visualization. Generative vision models have generated photorealistic imagery of faces~\cite{isola17pix2pix, wang18pix2pixhd}, animals~\cite{Zhu_2017b, brock2018large}, street-level flood imagery~\cite{Schmidt_2019}, and satellite observations~\cite{requena2019predicting, Fruhstuck_2019, Mohandoss_2020, Singh_2018, Audebert_2018}. Synthetic satellite imagery, however, needs to be explainable~\cite{Barredo20, ali2023explainable} and trustworthy \cite{diaz2023connecting}. 
Many complementary approaches exist to increase the trustworthiness of generative vision models, including interpretable networks~\cite{bau20lotteryticket}, adversarial robustness~\cite{Madry_2018,santamaria20truebranch,Lutjens_2019}, or probabilistic predictions with uncertainty~\cite{Gneiting_2005, Lugmayr_2020,Lutjens_2018}. Here, we raise a new question of 'How can we increase trust in synthetic satellite imagery through physical-consistency?'. 

We define a synthetic image to be physically-consistent if the depictions in the image are consistent with the output of a physics-based model, as detailed in~\cref{sec:phys_con}. 
Our definition of physical-consistency relates to the field of physics-informed machine learning (ML) in which researchers find novel ways to embed physics domain knowledge into deep learning methods~\cite{karniadakis21piml,raissi18hiddenphysics,Brunton_2019,Rasp_2018, Lutjens_2020}. 
We considered various physics-informed ML methods to incorporate the physics of floods in a generative vision model as
inputs~\cite{Reichstein_2019}, constrained representation~\cite{lusch18koopman, greydanus19hamiltonian, bau20lotteryticket}, training loss~\cite{raissi19pinns}, hard output constraints~\cite{Mohan_2020,donti21dc3,harder22downscaling}, or evaluation function~\cite{lesort19deep}. 
We also considered to embed a generative vision model into a set of physics equations, specifically, in the differential equations of floods as learned parameters~\cite{Garcia_2006, raissi19pinns}, dynamics~\cite{Chen_2018}, residual~\cite{karpatne17pgnn, yuval21stable}, differential operator~\cite{raissi18hiddenphysics, long19pdenet2}, or solution~\cite{raissi19pinns}. Finally, we decided to incorporate physics as input and evaluation functions. Specifically, we use a 1-channel flood mask, that represents the projections of a physics-based flood model, and 3-channel satellite imagery as inputs to a deep generative vision model and evaluate the intersection over union (IoU) of the generated image and flood input, as detailed in~\cref{sec:eie_methods}. 

We call our method \eie/ \textit{(EIE)} and associate a novel dataset in segmentation-guided image-to-image translation with it, as detailed in~\cref{sec:eie_data}. We show that our method outperforms a physics-unconditioned baseline model in~\cref{sec:eie_results_main} and discuss generalization across space, remote sensing instruments, and other climate events (reforestation and Arctic sea ice melt) in ~\cref{sec:eie_generalization_results}.

Our work makes the following contributions: \begin{itemize}
    \item A novel framework to measure physical-consistency in synthetic satellite imagery,
    \item the first physically-consistent and photorealistic visualization of flood risks as satellite imagery,
    \item an open-source dataset with over $30k$ labeled high-resolution image-triplets that can be used to study image-to-image translation in Earth observation.
\end{itemize} 



\section{Related Work}\label{sec:eie_related_works}
We generate physically-consistent visualizations of climate change-related changes in satellite imagery by formulating a semantic image synthesis task and applying deep generative vision models to solve it. 

\subsection{Generative vision modeling.}
We formulate the generation of satellite imagery as an image-to-image (im2im) translation problem: learn a mapping from satellite image and segmentation mask to another satellite image~\cite{isola17pix2pix}. Deep generative vision models have been most successful at solving im2im problems~\cite{isola17pix2pix}. Generative adversarial networks (GANs) have been successfully used in semantic image synthesis, a subproblem within im2im, to generate photorealistic street scenery from semantic segmentation masks: DCGAN~\cite{Radford_2016}, Pix2pixHD~\cite{wang18pix2pixhd}, DRPAN~\cite{wang18pix2pixhd}, SPADE~\cite{Park_2019}, or OASIS~\cite{Schonfeld_2021}. 
Similarly, probabilistic normalizing flows (NFs) ~\cite{rezende15normflows, Lugmayr_2020}, variational autoencoders (VAEs) ~\cite{Kingma_2013, Zhu_2017}, autoregressive models~\cite{parmar18autoregressive}, and diffusion-based models (\cite{saharia21palette} and~\cref{fig:dalle2}) have been adapted to im2im translation. Our use-case requires a deterministic semantic image synthesis model that is capable of generating realistic high-resolution (i.e., 1024x1024px) images. We decided to focus on GANs, because VAEs generate less realistic images (\cite{Dosovitskiy_2016, Zhu_2017} and \cref{fig:results_comparison_imagery}). NFs could likely capture the distribution of possible images more accurately, but our use-case of engaging visualizationsis sufficently captured with a single deterministic high-resolution image and NFs typically require specialized model architectures that can become computationally expensive. Diffusion-based models show comparable performance to GANs in generating synthetic satellite imagery~\cite{wolters23zooming}. However, diffusion-models are currently computationally too expensive to train at 1024x1024px resolution, can still be outperformed by GANs in similar tasks~\cite{wolters23zooming}, and feature stochasticity that is not needed for this task. 
So, we decided to extend the high-resolution semantic image synthesis model, pix2pixHD~\cite{wang18pix2pixhd}, to take in 4-channel images that include physical information and to generate satellite imagery that is both photorealistic and physically-consistent. 


\subsection{Climate change visualization tools}
Visualizations of climate change are commonly used in policy making and community discussions on climate adaptation~\cite{Sheppard_2012, Cohen_2012}.
Landscape or 'street-view' visualizations are used to raise environmental awareness in the general public or policy~\cite{Sheppard_2005,Schmidt_2019}, because they can convey the impacts of climate change, such as rising sea levels or coastal floods, in a compelling and engaging manner (\cite{Sheppard_2005},~\cref{fig:related_works}b). 
Most landscape visualizations, however, are limited to regional information~\cite{Strauss_2015}. Additionally, most landscape visualizations require expensive physics-based renderings and/or  high-resolution digital elevation models~\cite{Strauss_2015}. 
Alternative visualization tools of coastal floods or sea-level rise are color-coded maps, such as~\cite{NOAA_2020, NoaaSlosh_20, ClimateCentral_18}. Color-coded maps convey the flood extent on a city-wide scale, but are less engaging than a photorealistic image~\cite{Sheppard_2012}. We are generating compelling visualizations of climate change-related events as satellite imagery to aid in policy and community discussions on climate adaptation.

\section{Materials and Methods}\label{sec:eie_methods}

To synthesize satellite imagery that depicts floods, we have trained a deep generative vision model. Specifically, we trained a pix2pixHD GAN~\cite{wang18pix2pixhd} to translate a \textit{pre-flood} satellite image and a corresponding \textit{flood mask} segmentation map to a \textit{post-flood} satellite image that depicts the flood, as shown in~\cref{fig:eie_model_architecture_flood}. The mapping dimensions are: $[(1024,1024,3),(1024,1024,1)] \rightarrow (1024,1024,3)$. 
First, the model is trained and evaluated with flood masks that are derived from the post-flood images via a segmentation model (see~\cref{sec:flood_seg}) and, then, during inference the model uses flood masks from a physics-based flood model.
We train and evaluate the flood visualization and segmentation models on multiple datasets with a total of over $10k$ HD image-triplets. Our experimental setup evaluates if the model can visualize the precise flood extent as it would be simulated by a flood model --  the GAN model is not intended or evaluated to downscale or correct biases inherent in flood models.
We discuss an extension to generate imagery of reforestation and towards visualizing Arctic sea ice melt. We describe our datasets in~\cref{tab:data_overview} and~\cref{sec:eie_data}, the model architecture in~\cref{sec:model_architecture}, and we define physical-consistency in~\cref{sec:phys_con}.

\subsection{Data Overview.}\label{sec:eie_data}
\begin{table*}[t]
\caption[Data Overview]{\textbf{Data Overview.} We created nine datasets to study image-to-image (im2im) translation and segmentation (seg) for flood, reforestation, and sea ice melt events. In total, the datasets contain $\approx 90GB$ or $32k$ HD image-pairs or -triplets.}
    \centering
    \subfloat{
  \includegraphics [trim=0 0 0 0, clip, width=0.98\linewidth, angle = 0]{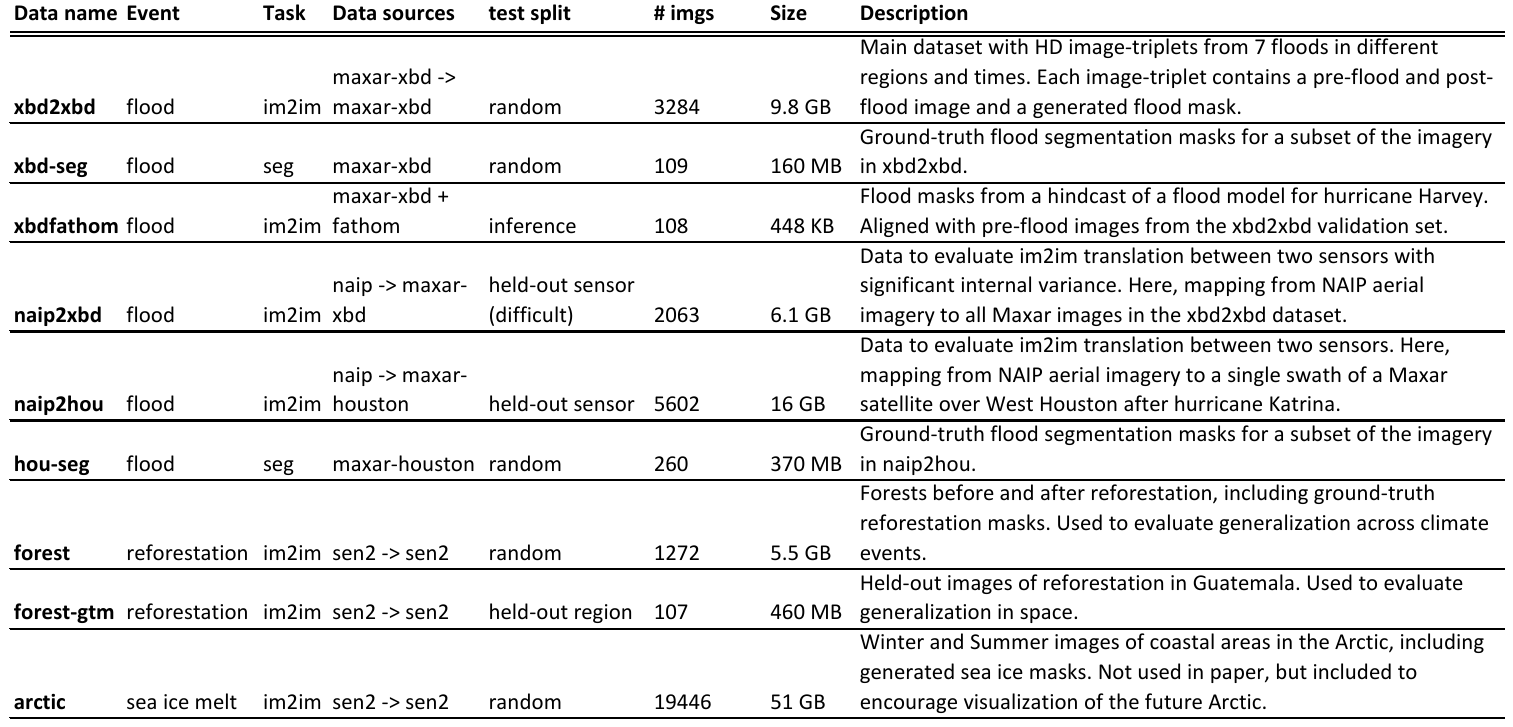}}
\label{tab:data_overview} 
\end{table*}

As part of this study, we have created nine open-source datasets in flooding, reforestation, and sea ice melt, which are summarized in~\cref{tab:data_overview} and detailed in Appendix~\ref{sec:appendix_dataset}. The datasets are formatted as image-triplets for creating an image-to-image (im2im) translation model or as image-pairs for creating a segmentation (seg) model. In total, the data is approximately $90 \unit{GB}$ or $32\unit{k}$ sets of images.

\subsubsection{Main flood dataset}
In our main dataset, \textbf{xbd2xbd}, we have assembled 3284 image-triplets from seven flood events in different regions and time, as detailed in Appendix~\ref{sec:appendix_dataset}. Our method requires a large collection of \textit{post-event} images that depict how climate events impact a landscape. However, these post-event images are usually challenging to acquire. In the case of flooding, obtaining post-flood images that display standing water is hindered by cloud-cover, time of standing flood, satellite revisit rate, atmospheric noise, and cost of high-resolution imagery. For the xbd2xbd dataset, we have sourced it by selecting flood-related data from the xBD xview2 dataset~\cite{Gupta_2019}. Here, each post-flood image is already paired with a pre-flood image, which was taken by the same satellite constellation over the same region before the flood struck. The pre- and post-flood data have HD $1024{\times1024}\unit{px/img}$ resolution with ground-sample distance ${\sim}.5\unit{m/px}$, are RGB, and were taken by the Maxar DigitalGlobe satellites.

To add the layer of physical consistency, we associate the pre- and post-flood pairs with a binary low- (${\sim}30\unit{m/px}$) or high-resolution (${\sim}.5\unit{m/px}$) flood mask. 
Because flooding hindcasts typically do not exactly match the observed flood extent (see e.g.,~\cite{wing19fathomharvey} for hurricane Harvey), we derive the flood mask inputs that are used to train the im2im model from the post-flood images using a separate flood segmentation model. To train this flood segmentation model, we have created the \textbf{xbd-seg} dataset which contains $109$ pairs of post-flood images and corresponding hand-labelled flood segmentation masks.

\subsubsection{Auxiliary datasets}
After training and evaluating the im2im model on masks of the observed flood extent, we test if it can also visualize predictions from a physics-based flood model using the \textbf{xbdfathom} dataset. The xbdfathom dataset pairs the pre-flood images in the xbd2xbd hurricane Harvey validation set with ${\sim}30\unit{m/px}$ flood masks from the Fathom-US hydraulic model framework hindcast, as detailed in Appendix~\ref{sec:appendix_data_xbdfathom}. 
The datasets, \textbf{naip2xbd}, \textbf{naip2hou}, and \textbf{hou-seg}, are used in~\cref{sec:gen_loc} to test if our model could visualize flooding by using NAIP aerial imagery as input, which would be available across the full U.S. East Coast. The datasets, \textbf{forest} and \textbf{forest-gtm}, are used to study reforestation visualizations in~\cref{sec:gen_forest} and described in Appendix~\ref{sec:appendix_data_forest}. Finally, we created the \textbf{arctic} dataset in~\ref{sec:appendix_data_arctic}, which we did not use in model training, but publish to facilitate extensions of our method for visualizing sea ice melt.

\begin{figure}[t]
  \vspace{-.2in}
  \centering
      \includegraphics [trim=0 .3 0 0, clip, width=1.\columnwidth, angle = 0]{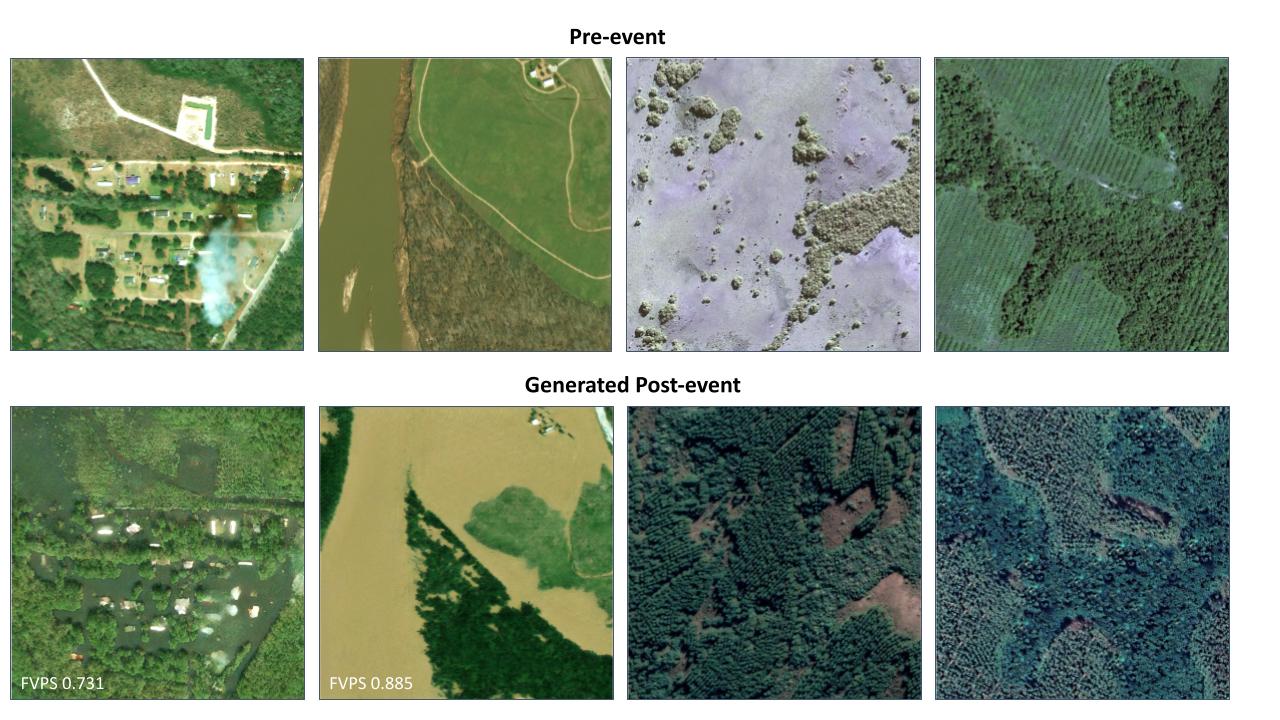}
      \vspace{-.2in}
\caption[Overview of synthesized imagery]{Our model, titled the \eie/, visualizes how flooding (left) or reforestation (right) would impact the landscape as seen from space.}\label{fig:results_overview} 
\end{figure}

\subsection{Model architecture.}\label{sec:model_architecture}
The central model of our pipeline is a generative vision model that learns the physically-conditioned image-to-image transformation from pre-flood image to post-flood image. We leveraged the existing implementation of pix2pixHD~\cite{wang18pix2pixhd}. Pix2pixHD is a state-of-the-art semantic image synthesis model that uses multi-scale generator and discriminator architectures to generate high-resolution imagery and we refer to the original paper~\cite{wang18pix2pixhd} for details on the model architectures. We extended the input dimensions to $1024{\times}1024{\times}4$ to incorporate the focus area mask. We experimented with different parameters (see~\cref{sec:appendix_experiments}), but eventually decided to use the same architecture and hyperparameters as in~\cite{wang18pix2pixhd}. We trained the model from scratch on each dataset. For the xbd2xbd dataset, training took $200$ epochs in ${\sim}7\unit{hrs}$ on $8{\times}\unit{V100}$ Google Cloud GPUs. The resulting pipeline is modular, such that it can be repurposed for visualizing other climate impacts.

\subsection{Trust in flood images through physical-consistency.}\label{sec:phys_con}
We define a \textit{physically-consistent} model as one that fulfills laws of physics, such as, conservation of momentum, mass, and energy~\cite{Brunton_2019}. For example, most coastal flood models consist of numerical solvers that resolve the conservation equations to generate flood extent predictions~\cite{Jelesnianski_92}.
Here, we consider a flood image to be physically-consistent if it depicts the predictions of a physically-consistent model. 

Specifically, we define our generated satellite imagery, $I_G~\in~\mathcal I = [0,1]^{w\times h\times c}$ with width, $w=1024$, height, $h=1024$, and number of channels, $c=3$, to be physically-consistent if it depicts the same flood extent as the binary flood mask, $F\in \mathcal F = \{0;1\}^{w\times h}$. 
We implemented a flood segmentation model, $m_{\text{seg}}:\mathcal I \rightarrow \mathcal F$, to measure the depicted flood extent in the generated image. If the flood extent of a generated image and the coastal flood model match within a margin, the image is in the set of physically-consistent images, i.e,. 
$\mathcal I_{\text{phys}} = \{I_G \in \mathcal I:\; \rvert\rvert m_{\text{seg}}(I_G) - F\lvert\lvert~<~\epsilon\}$. The generated image is considered photorealistic, if it is contained in the manifold of naturally possible satellite images, $\mathcal I_{\text{photo}} \subset \mathcal I$. Hence, we are looking for a conditional image generation function, $g$, that generates an image that is both, physically-consistent and photorealistic, i.e, $g :\mathcal I_\text{photo} \times \mathcal F \rightarrow \mathcal I_\text{photo} \cap \mathcal I_\text{phys}$. Here, we condition the GAN on the flood mask, $F$, and use a custom evaluation function to identify the generation function, $g$.


\begin{figure*}[t]
  \centering
  \begin{subfigure}{1.\textwidth}
      \centering
      \includegraphics [trim=0 0 0 0, clip, width=1.\textwidth, angle = 0]{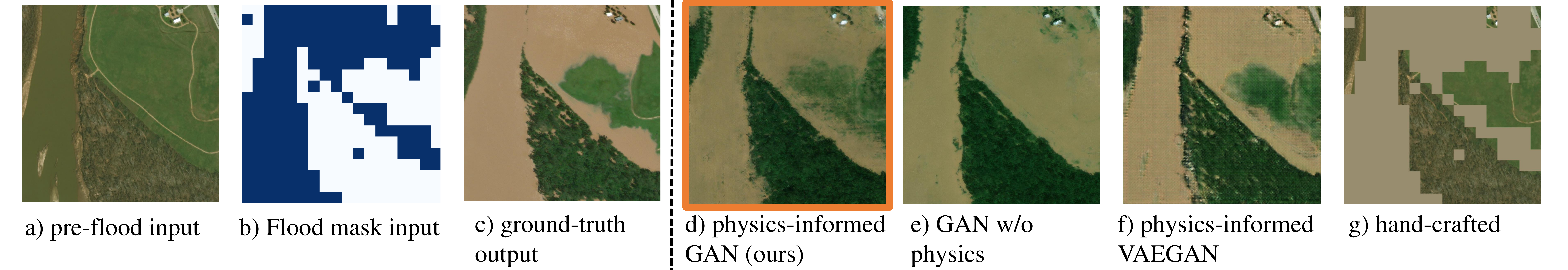}
      \includegraphics [trim=0 0 0 0, clip, width=0.85\textwidth, angle = 0]{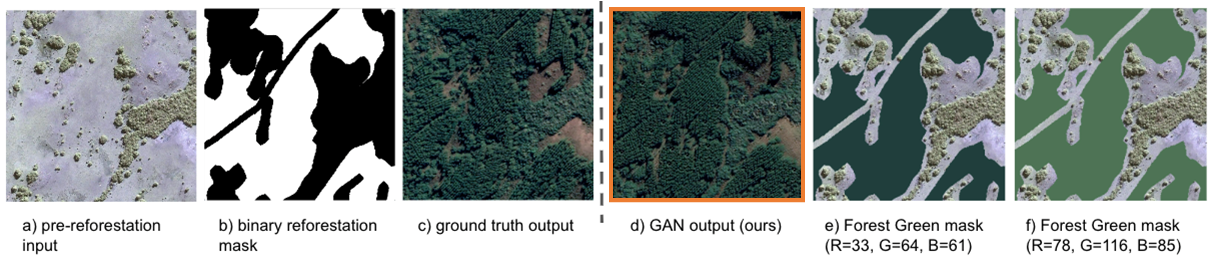}
  \end{subfigure}
\caption[Qualitative comparison of generated imagery]{\textbf{Top: Flooding.} The proposed physics-informed GAN (ours, based on~\cite{wang18pix2pixhd}) generates post-flood images, (d), from a pre-flood image and flood mask, (a,b). The model outperforms others, (e,f,g), in either physical-consistency or photorealism. The baseline GAN ablates the flood mask and synthesizes a fully-flooded image, (e), rendering the model physically-inconsistent. A VAE-based model~\cite{Zhu_2017}, creates glitchy imagery, (f), (zoom in). A handcrafted baseline model as used in common visualization tools~\cite{ClimateCentral_18, NOAA_2020}, visualizes the correct flood extent, but is pixelated and lacks photorealism, (g).

\textbf{Bottom: Reforestation.} The proposed reforestation mask + pix2pixHD~\cite{wang18pix2pixhd} GAN, (d), generates photorealistic reforestation imagery from the inputs, (a,b), outperforming handcrafted baseline models (e,f). 
} 
\label{fig:results_comparison_imagery} 
\vspace{-.1in}
\end{figure*}

\section{Results}\label{sec:eie_results}
In~\cref{sec:results_eval_metrics}, we define the evaluation metrics.~\Cref{sec:eie_results_main} analyses the physical-consistency and visual quality of the generated flood imagery on the xbd2xbd dataset and~\cref{sec:flood_seg} evaluates the underlying flood segmentation model.
In~\cref{sec:gen_physics}, we query the trained im2im model with flood masks from a physics-based flood model for selected locations in Houston, TX. In~\cref{sec:gen_loc}, we evaluate which steps would be necessary to generate a flood visualization layer across the full U.S. East Coast. In~\cref{sec:gen_forest} and~\cref{sec:gen_arctic}, we extend the model to visualize reforestation, and describe a dataset for future extensions towards visualizing Arctic sea ice melting.

\subsection{Evaluation Metrics.}\label{sec:results_eval_metrics}
Evaluating synthetic imagery is difficult~\cite{xu2018empirical,Borji_2019}. 
Most evaluation metrics measure photorealism or sample diversity~\cite{Borji_2019}, but not physical consistency~\cite{ravi2019adversarial} (see, e.g., SSIM~\cite{Wang_2004}, MMD~\cite{Bounliphone_2016}, IS~\cite{salimans2016improved}, MS~\cite{Che_2017}
, FID~\cite{heusel2017gans, zhou20establishing}, or LPIPS~\cite{zhang18lpips}). 

To evaluate physical consistency we propose using the intersection over union (IoU) between water in the generated post-flood image and water in the ground-truth flood mask. This method relies on flood masks, but because there are no publicly available flood segmentation models for high-resolution visual satellite imagery, we trained our own model (\cref{sec:flood_seg}).
This segmentation model created flood masks of the generated and ground-truth post-flood image which allowed us to measure the overlap of water in between both (best IoU is $1$, lowest is $0$). 

To evaluate photorealism, we used the commonly used perceptual similarity metric Learned Perceptual Image Patch Similarity (LPIPS)~\cite{zhang18lpips}. LPIPS computes the feature vectors (of an ImageNet-pretrained AlexNet CNN architecture) of the generated and ground-truth post-flood image and returns the mean-squared error between the two feature vectors (best LPIPS is $0$, worst is $1$).

Because the joint optimization over two metrics poses a non-trivial hyperparameter optimization problem, we propose to combine the evaluation of physical consistency (IoU) and photorealism (LPIPS) in a new metric (FVPS), called Flood Visualization Plausibility Score (FVPS). 
The FVPS is the harmonic mean over the submetrics, IoU and $(1-\text{LPIPS})$, that are both $[0,1]$-bounded.
Due to the properties of the harmonic mean, the FVPS is $0$ if any of the submetrics is $0$; the best FVPS is $1$. In other words, the FVPS is only $1$ if the imagery is both photorealistic and physically-consistent. A small number, $\epsilon$, is added for numerical stability. 
\begin{equation}
    \text{FVPS} = \frac{2}{\frac{1}{\text{IoU} + \epsilon} + \frac{1}{1-\text{LPIPS} + \epsilon}}.\label{eq:fvps}
    \vspace{.1in}
\end{equation} 

\subsection{Physical-consistency and photorealism.}\label{sec:eie_results_main}
We train and evaluate our physics-informed GAN on the xbd2xbd dataset, as detailed in~\cref{sec:eie_methods}. The dataset contains over $3k$ HD image-triplets from floods in multiple regions and times. We used all regions during training and evaluated on randomly held-out images of two regions with most visible floods. We evaluate all models on inference with high-res. (${\sim}.5\unit{m/px}$) and coarse-grained low-res. (${\sim}30\unit{m/px}$) flood masks to imitate flood model predictions at different resolutions. We compare our model against a GAN with the same model architecture, but an ablated flood mask (baseline GAN) in~\cref{sec:results_gan_ablation}, a photoshopped baseline with perfect flood extent in~\cref{sec:results_handcrafted_baseline}, and a VAE-based model in~\cref{sec:results_vae_gan}. Overall, we find that our model is on-par with the VAE and outperforms the GAN and photoshopped baseline in terms of physical-consistency or photorealism, respectively.

\subsubsection{A GAN without physics information generates photorealistic, but non physically-consistent imagery.}\label{sec:results_gan_ablation} 
The baseline GAN uses the default pix2pixHD~\cite{wang18pix2pixhd} with the same architecture and hyperparameters as the physics-informed GAN, but only uses the pre-flood image and not the flood mask as input. The baseline GAN visualized floods at locations where there was no flood according to the flood mask, for example, in \Cref{fig:results_comparison_imagery}e-top vs. ~\cref{fig:results_comparison_imagery}b-top. In practice, it would be dangerous to operationalize a visualization with false-positive flood predictions. And, the inaccurately modeled flood extent illustrates the importance of measuring physical-consistency, as defined in~\cref{sec:phys_con}.
In our case, we measure physical-consistency via the IoU of the flood masks and, across the high-res. validation set (in Appendix~\ref{sec:appendix_dataset}), the baseline GAN has a lower IoU ($0.226$) in comparison to the physics-informed GAN ($0.502$). 
Despite the photorealism of the baseline GAN ($\unit{LPIPS}=0.293$ vs. $0.265$), the physical-inconsistency renders the model non-trustworthy as confirmed by the low FVPS ($0.275$).

\subsubsection{A handcrafted baseline model generates physically-consistent but not photorealistic imagery.}\label{sec:results_handcrafted_baseline} Similar to common flood visualization tools~\cite{ClimateCentral_18}, the handcrafted model overlays the flood mask input as a hand-picked flood brown (\#998d6f) onto the pre-flood image, as shown in~\cref{fig:results_comparison_imagery}g-top. On the low-res. validation set, this model should have a perfect IoU ($1$) by construction, but our GAN-based flood segmentation model struggles with hard boundaries and measures an acceptable IoU ($0.361$). Qualitatively, the handcrafted baseline looks pixelated with flood masks at the resolution of a physical model~\cite{NoaaSlosh_20} of $30\unit{m/px}$ and does not account for varying flood color between events (see~\cref{fig:results_overview}). The high LPIPS ($0.415$) indicates quantitatively that the handcrafted visualizations are less photorealistic then the physics-informed GAN ($0.283$). Despite the lower photorealism, we believe that a handcrafted baseline is a reasonable alternative to generative vision models considering the benefit-cost ratio. In practice, the handcrafted baseline's main benefit would be the perfect IoU and the GAN's would be the interpolation of low res. flood masks onto the imagery's resolution, which can become especially relevant in densely populated areas.

\subsubsection{The proposed physics-informed GAN generates physically-consistent and photorealistic imagery.}\label{sec:results_vae_gan}
To create the physics-informed GAN, we trained a pix2pixHD~\cite{wang18pix2pixhd} on the xbd2xbd data, as detailed in~\cref{sec:eie_methods} and Appendix~\ref{sec:appendix_experiments}. This model successfully learned how to synthesize photorealistic post-flood images from a pre-flood image and a flood mask, as shown in~\cref{fig:results_overview}-left. The 
model improves over all other models either in terms of IoU, LPIPS, or FVPS (\cref{tab:results_comparison_table}). The learned image transformation
``in-paints`` the flood mask in the correct flood colors and displays an average flood \textit{height} that does not cover structures (e.g., buildings, trees), as shown in $64$ randomly sampled test images in~\cref{fig:grid_of_gen_ims}. Occasionally, city-scenes show scratch patterns, e.g.,~\cref{fig:grid_of_gen_ims} (top-left). This could be explained by the unmodeled variance in off-nadir angle, sun inclination, GPS calibration, color calibration, atmospheric noise, dynamic objects (cars), or flood impacts, which is partially addressed in our generalization experiments~\cref{sec:gen_loc}.
We also train a VAE-based model with GAN components on the same in-/outputs than the physics-informed GAN. Specifically, we use a VAEGAN called BicyleGAN~\cite{Zhu_2017}, which has the potential to create ensemble forecasts over the unmodeled flood impacts, such as the probability of destroyed buildings. The generated VAEGAN images look smeared~\cref{fig:results_comparison_imagery}f-top and the LPIPS ($0.449$) is worse than for the physics-informed GAN ($0.265$) on the high res. validation set.

\begin{table*}[tp]
\centering
\vspace{0.1in}
\caption[Benchmarking physics-informed GAN on flood visualizations]{\textbf{We evaluate photorealism (LPIPS) and physical consistency (IoU)}, to compare our physics-informed GAN with three baselines: a baseline GAN with ablated flood mask; a physics-informed VAEGAN; and a handcrafted baseline. The proposed Flood Visualization Plausibility Score (FVPS) trades-off IoU and LPIPS as a harmonic mean and highlights the performance differences between the GAN with and without physics on low-resolution flood mask inputs. The $\downarrow$ ($\uparrow$) arrows indicate that a lower (higher) score is better, respectively.}
\begin{tabular}{c|c c c c c c}  
  & \thead{LPIPS $\downarrow$ \\ high res.} & \thead{LPIPS $\downarrow$ \\ low res.} & \thead{IoU $\uparrow$\\ high res.} & \thead{IoU $\uparrow$\\ low res.}  & \thead{FVPS $\uparrow$ \\ high res.}  & \thead{FVPS $\uparrow$\\ low res.} \\ [0.5ex]
 \hline\hline 
 \textbf{GAN w/ phys. (ours)} & \textbf{0.265} & \textbf{0.283} & \textbf{0.502} & \textbf{0.365} & \textbf{0.533} & \textbf{0.408} \\ \hline
 GAN w/o phys. & \textbf{0.293} & \textbf{0.293} & 0.226 & 0.226 & 0.275 & 0.275 \\[0.5ex] \hline
 VAEGAN w/ phys. & 0.449 & - & \textbf{0.468} & - & 0.437 & - \\ \hline
 Handcrafted baseline & 0.399 & 0.415 & \textbf{0.470}  & \textbf{0.361} & 0.411 & \textbf{0.359} \\[0.5ex]
\end{tabular}
\label{tab:results_comparison_table}
\vspace{-.1in}
\end{table*}

\subsection{Flood segmentation model.}\label{sec:flood_seg}
Our approach requires a flood segmentation model to generate ground-truth flood masks and evaluate the generated post-flood images, but there does not exist any open-source model that segments floods in high-resolution ($<$1m/px) satellite imagery. Hence, we labelled our own dataset of 109 HD image-pairs, xbd-seg, and trained a flood segmentation model. Our model is based on pix2pix~\cite{isola17pix2pix} with a custom loss function and validated with 5-fold cross validation due to the small data size. The model is trained and evaluated at ${\sim}.5\unit{m/px}$ and flood masks at ${\sim}30\unit{m/px}$ are derived via nearest neighbor downsampling of the model outputs. 
The segmentation model at ${\sim}.5\unit{m/px}$ has a mean IoU of $0.343$ which matches the expected performance of pix2pix~\cite{isola17pix2pix} and is sufficient for our task. While one might expect an IoU closer to $1.0$, many of our ground-truth flood masks have $<5\%$ of positive labels which skews the IoU towards zero if not perfectly predicted (see, e.g.,~\cref{fig:results_flood_seg}-bottom-left). Our segmentation model is detailed in Appendix~\ref{sec:appendix_flood_seg} and the xbd-seg data in Appendix~\ref{sec:appendix_data_xbd_seg}.

\subsection{Generalization performance.}\label{sec:eie_generalization_results}
Our project aims to create satellite imagery that visualizes climate phenomena across the globe. 
So far, however, we have only evaluated our model on one climate phenomenon (floods), one remote sensing instrument (Maxar satellite imagery), a few selected locations, and observation-derived flood masks. 
To extend our model, we first test if the im2im model can be applied to flood masks from a physics-based flood model. 

\subsubsection{Inference with physics-based flood mask inputs.}\label{sec:gen_physics}

We query the physics-informed GAN -- trained on xbd2xbd -- for the pre-flood imagery from hurricane Harvey and flood masks in xbd2xbd and xbdfathom. To qualitatively analyse the results, we first plot the predictions (3rd col.) when using the low-res. flood mask inputs (2nd col.) from the xbd2xbd validation dataset, in Appendix~\ref{fig:xbdfathom_generated_images}. Then, we use the same model and pre-flood image (1st col.), but use the flood masks from the Fathom Global hydraulic model (6th col.) in xbdfathom as input and plot the predictions in the 5th column.

For both datasets, the generated imagery seems to contain brown-colored floods in areas matching the flood mask inputs. Especially, the images in the 2nd row illustrate the capacity of the im2im model to visualize a different flood extent depending on the input mask. The average IoU across all images in xbdfathom dataset is $0.398$ which is on a similar scale as the IoU on the low-res. xbd2xbd data ($0.365$) in~\ref{tab:results_comparison_table}. Limitations of the im2im model are visible in forests occasionally being over-painted with a flood brown or houses that appear smeared and are discussed in Appendix~\ref{sec:appendix_results}. Overall, these results indicate that the model can generalize from being trained with observation-derived flood masks to being tested with physics-simulated flood masks, assuming that the distribution of pre-flood images is held constant.

\subsubsection{Generalization across location and remote sensing instruments.}\label{sec:gen_loc}
In order to create a visualization layer of flood hazards across the U.S. East Coast, we would need, among others: flood mask inputs across the U.S. East Coast, pre-flood images across the U.S. East Coast, and a model that has been validated for these data streams. We also note that our work only evaluates the im2im model on a tile-by-tile basis and stitching multiple tiles into a seamless large-scale tif entails a set of challenges that are beyond the scope of this work~\cite{li19mosaicking,fruhstuck19tilegan}.

The flood mask inputs are relatively easy to access, for example, by using a NWS SLOSH National Storm Surge Hazard Map for a hurricane category X storm [5] as discussed in Appendix E. Because this layer is only available at 30m/px, we tested our model’s performance on low-res. flood masks
in Section IV-B.

The flood mask inputs at $\approx30\unit{m/px}$ resolution could be accessed for research purposes, for example, from the Fathom Global flood inundation model which was run for multiple climate scenarios~\cite{wing23fathomglobalssp}. Alternatively, the National Weather Service publishes a $30\unit{m/px}$ National Storm Surge Hazard Map for hurricanes of different categories~\cite{NoaaSlosh_20}. As~\cref{sec:gen_physics} showed that $30\unit{m/px}$ physics-based flood masks can be used as inputs, we consecutively evaluate if different pre-flood images can be used as inputs.

The pre-flood images in xbd2xbd are from Maxar, which is not freely available across the full U.S. East Coast. Hence, we downloaded pre-flood images from the open-access U.S.-wide mosaic of $1.0m/px$ visual aerial imagery from the 2019 National Agriculture Imagery Program (NAIP)~\cite{NAIP_2019}. For the new dataset, naip2xbd, we pair the pre-flood NAIP images with the post-flood Maxar images and flood masks from xbd2xbd. During the pairing process, we orthorectified the Maxar images to match the NAIP images. 

The im2im task from NAIP to Maxar imagery in naip2xbd is significantly more challenging than the Maxar to Maxar task in xbd2xbd, because the learned image-transformation needs to account for the sources of variance in differing remote sensing instruments. These are, for example, resolution, atmospheric noise, color calibration, inclination angle, and other factors. We first applied the physics-informed GAN from xbd2xbd without fine-tuning on the new naip2xbd data, but saw that it generates unintelligible imagery in~\cref{fig:houston_west}-top. After re-training the same model from scratch on the naip2xbd data, the image quality was still relatively poor (not shown). We believe that the variance within the xbd2xbd post-flood images data was too large to learn a mapping from NAIP to Maxar imagery with only 2063 HD images in naip2xbd.

To reduce the learning task complexity, we created the naip2hou dataset, for which we sourced post-flood image tiles from a single open-access Maxar satellite pass over West Houston, TX on 8/31/2017, post-hurricane Harvey~\cite{Harvey_2017}. 
To re-run our pipeline, we labelled an additional $260$ flood segmentation masks (hou-seg, taking $\sim 20\unit{hrs}$), retrained the flood segmentation model, and generated flood masks for the naip2hou dataset, as shown in~\cref{fig:results_flood_seg}. The naip2hou dataset contains $5602$ HD image-triplets. Then, we re-trained the physics-informed GAN from scratch on the new dataset for 15 epochs in $\sim 5\unit{hrs}$ on $1\times V 100$ GPU. We use a random 80-20 train-val split.

The resulting model for naip2hou has acceptable performance, illustrated in~\cref{fig:houston_west}-top. It does not outperform a handcrafted baseline model in physical-consistency (not shown), but outperforms it in photorealism (LPIPS=$0.369$ vs. $0.465$). This indicates that im2im translation across remote sensing instruments is feasible and a visualization of flood hazards along the U.S. East Coast could be realized in follow-up work.

With xbd2xbd and naip2hou, we created a dataset of a combined $8886$ clean image-triplets that we are releasing as the flood-section of our open-source dataset to study segmentation guided image-to-image (im2im) translation in Earth observation. Further, albeit small and geographically biased, the xbd-seg and hou-seg data with a combined 369 HD image-pairs are the first open-source datasets for flood segmentation on visual high-resolution ($<1m/px$) satellite imagery, to the extent of the authors' knowledge, and will be made available as part of the dataset. 


\subsubsection{Generalization across different climate phenomena -- visualizing reforestation}\label{sec:gen_forest}
Visualizing negative climate impacts such as flooding might evoke anger, fear, or guilt in some viewers. These emotions can encourage pro-environmental behavior~\cite{thomas09anger,slovic04fear,bamberg07guilt}, but also cause inaction from feeling overwhelmed~\cite{oneill09overwhelming} and hope is needed to maintain environmental engagement~\cite{ojala12hope}. Here we extend the \eie/ to visualize the impact of positive hopeful actions, specifically, reforestation. This visualization has already and can be used to encourage policymakers, carbon finance investors~\cite{reiersen22reforestree,lutjens19forests}, or landowners to increase reforestation efforts.  

To synthesize satellite imagery of reforested land, the \eie/ uses an image of a deforested area along with a binary mask of where trees will be planted as input, as illustrated in~\cref{fig:model_architecture_forest}. 
To train and evaluate the model we collected the \textbf{forest} dataset, which contains image-triplets of an RGB pre-reforestation satellite image, a binary reforestation area mask (1=reforestation), and an RGB post-reforestation satellite image. We assembled a total of 1026 train and 246 test high-resolution (50cm/px) 1024x1024 image-triplets via Google Earth Pro (Map data: Google, Maxar Technologies, CNS/Airbus). The dataset spans four different countries: Uruguay, Sierra Leone, Peru, and Mexico, as detailed in Appendix~\ref{sec:appendix_data_forest}. 

We trained the generative vision model that performed best on floods, pix2pixHD, using several augmentation techniques and the default~\cite{wang18pix2pixhd} hyperparameters, as detailed in Appendix~\ref{sec:appendix_augmentations}. We evaluated the model on a random validation split on \textbf{forest} and a spatial split testing on 107 held-out images, \textbf{forest-gtm}, from Guatemala. We evaluate LPIPS and compare the GAN to a baseline model that applies uniformly colored masks. 

\Cref{fig:results_comparison_imagery}d)-bottom shows how our model generates photorealistic visualizations of reforestation projects. The generated imagery looks more realistic than handcrafted baseline models (e,f), where the reforested area pixels are set to a mean forest color. Our quantitative analysis in~\cref{tab:results_comparison_table_forest} confirms that our model outperforms the baselines in both, a random and spatial split.We plot an additional random selection of generated images with their respective inputs in~\cref{fig:results_forest_random_10}.

\begin{table*}[tp]
\centering
\vspace{0.1in}
\caption[Evaluating Earth Intelligence Engine on reforestation]{\textbf{Reforestation accuracy}. Our model quantitatively outperforms two baseline models that apply a color mask in a random and spatial split.}
\begin{tabular}{c|c c}  
  & \thead{LPIPS $\downarrow$ \\ random split} & \thead{LPIPS $\downarrow$ \\ spatial split} \\ [0.5ex]
 \hline\hline 
 \textbf{GAN (ours)} & \textbf{0.503} & \textbf{0.574} \\ \hline
 Green mask (RGB=33,64,61) & 0.794 & 0.848 \\[0.5ex] \hline
 Green mask (RGB=78,116,85) & 0.845 & 0.957 \\[0.5ex] 
\end{tabular}
\label{tab:results_comparison_table_forest}
\vspace{-.1in}\end{table*}

\subsubsection{Visualizing Arctic sea ice melt}\label{sec:gen_arctic} 
The retreat of the Summer Arctic sea ice extent is one of the most important and imminent consequences of climate change~\cite{IPCC_2018}. However, visualizations of melting Arctic sea ice are limited to physics-based renderings, such as~\cite{NasaIce_2020}. There also does not exist publicly-available daily high-resolution (less than 500m) visual satellite imagery due to satellite revisit rate and cloud cover. To enable the extension of the \eie/ for visualizing Arctic sea ice melt, we publish the \textbf{arctic} dataset of $\approx 20k$ image triplets of Winter image, Summer image, and ice segmentation mask, as detailed in Appendix~\ref{sec:appendix_data_arctic}.


\section{Discussion}\label{sec:discussion}
We proposed a new pipeline to create synthetic and physically-consistent satellite images of future climate events using deep generative vision models. In this section, we discuss limitations and future work.
\subsubsection{Limitations.} 
First of all, satellite imagery is currently an internationally trusted source for analyses in deforestation, development, or military domains~\cite{Hansen_2013, Anderson_2017}. 
With the increased capability of data-generating models, more work is needed in the identification of and the education around misinformation and ethical and trustworthy AI~\cite{santamaria20truebranch,Barredo20}. We point out that our satellite imagery is synthetic, should only be used as a scientific communication aid to better explain our results to decision makers or the general public~\cite{Sheppard_2012}, and we take first steps towards guaranteeing trustworthiness in synthetic satellite imagery.

We had originally envisioned an operational coastal flood visualization across the U.S. East Coast, but discovered that the variance within and between remote sensing instruments is too large to develop an operational visualization as part of a single research paper. Our dataset is not small, as 3-5k HD image-triplets would equal 192-320k triplets at 128x128px resolution. However, our dataset contains spatial and temporal biases including a bias for U.S. areas and vegetation-filled areas. The latter likely contributes to our model rendering human-built structures, such as streets and out-of-distribution skyscrapers in~\cref{fig:grid_of_gen_ims} top-left, as smeared. Apart from the data limitation, smeared features are still a current concern in state-of-the-art GAN architectures~\cite{Schonfeld_2021} and generative vision model continue to struggle with the sharp lines in remote sensing imagery~\cref{fig:dalle2}. We overcame part of our data limitations by focusing our study on Houston, Tx and using several augmentation techniques, detailed in Appendix~\ref{sec:appendix_experiments}, but this work would likely still benefit from more diverse post-flood images from, e.g., the Maxar or NOAA data archives, or incorporating pre-trained geospatial embeddings~\cite{klemmer2023satclip,wolters23zooming}. 

\subsubsection{Future Work} 
We envision a tool that can visualize local climate impacts and adaptation techniques at the global scale.
By changing the input data, future work can visualize impacts of other well-modeled, visible, and climate-attributed events, including Arctic sea ice melt, hurricanes, wildfires, or droughts. Non-binary climate impacts, such as inundation height, or drought strength could be generated by replacing the binary flood mask with continuous model predictions. Opportunities are abundant for further work in visualizing our changing Earth. This work opens exciting possibilities in generating realistic and physically-consistent imagery
with the potential impact of improving climate mitigation and adaptation.
%



\section{Author Contributions}
Conceptualization, B.Lü. and B.Le.; methodology, B.Lü., B.Le., C.R.-M., F.C., N.D.-R., O.B., and A.L.; software, B.Lü. and M.M.-F.; validation, B.Lü.,A.P.,Y.G.,C.R.,and D.N.; formal analysis, B.Lü., B.Le., C.R.-M., F.C., N.D.-R., O.B., A.L., and M.M.-F.; investigation, B.Lü. and M.M.-F.,; data curation - floods, B.Lü., B.Le., C.R.-M., F.C., N.D.-R., O.B., A.L., and A.M.-P.; data curation - ice, A.S. and B.Lü.;data curation - forest, M.M.-F. and B. Lü.;writing—original draft preparation, B.Lü.; 
writing—review and editing, B.Lü., B.Le., C.R.-M., F.C., N.D.-R., O.B., A.L., A.S., and M.M.-F.; visualization, B.Lü.; supervision, B.Lü., D.N.; 
project administration, B.Lü. and B.Le.; funding acquisition, B.Lü., B.Le.,D.N. All authors have read and agreed to the published version of the manuscript.

BL\"u is leading the work and contributed to all sections. BLe started the work. BLe, CRM, FC, NDR, OB, AMP, and AL contributed to the flood experiments in III A-C. AS contributed to the ice experiment. MMF contributed to the reforestation experiment. AP, YG, CR, and DN advised the work.

\section{Funding}
This research was sponsored by the United States Air Force Research Laboratory and the United States Air Force Artificial Intelligence Accelerator and was accomplished under Cooperative Agreement Number FA8750-19-2-1000. The views and conclusions contained in this document are those of the authors and should not be interpreted as representing the official policies, either expressed or implied, of the United States Air Force or the U.S. Government. The U.S. Government is authorized to reproduce and distribute reprints for Government purposes notwithstanding any copyright notation herein. 

This research was partially conducted at the Frontier Development Lab (FDL), US. For the duration of FDL, the authors gratefully acknowledge support from the MIT Portugal Program, National Aeronautics and Space Administration (NASA), and Google Cloud.

\section{Data Availability Statement}
The code for this study is available at \href{https://github.com/blutjens/eie-earth-public}{github.com/blutjens/eie-earth-public}. The data has been published at 
\href{https://huggingface.co/datasets/blutjens/eie-earth-intelligence-engine}{huggingface.co/datasets/blutjens/eie-earth-intelligence-engine}.


\section{Acknowledgements}
We are very thankful for 
Leo Silverberg and Margaret Maynard-Reid for creating the demo at \href{https://climate-viz.github.io/} {https://climate-viz.github.io
}. We thank Ritwik Gupta for the continuous help in using the xBD dataset, Richard Strange for the help with cloud compute, Prof. Marco Tedesco for advise on the Arctic sea ice, Guy Schumann on flood modeling, Mark Veillette and Cait Crawford for technical direction, and James Parr, Leah Lovgren, Sara Jennings and Jodie Hughes for the organization of FDL and enabling these connections. We greatly appreciate the advise on decision-/policymaking in coastal climate adaptation by Derek Loftis, Sagy Cohen, Capt. John Radovan, Maya Nasr, and Janot Mendler de Suarez. Further, we greatly appreciate the feedback and direction from Esther Wolff, Hannah Munguia-Flores, Peter Morales, Nicholas Mehrle, Prof. Bistra Dilkina, Freddie Kalaitzis, Graham Mackintosh, Michael van Pohle, Gail M. Skofronick-Jackson, 
Tsengdar Lee, Madhulika Guhathakurta, Julien Cornebise, Maria Molina, Massy Mascaro, 
Scott Penberthy, John Karcz, Jack Kaye, Mich Lin, Campbell Watson, the FDL research community, and the anonymous reviewers. The authors also gratefully acknowledge support from Earthshot Labs for the reforestation visualization and greatly appreciate the feedback and help from Patrick Leung. We thank Phil Cherner and the members of the Harvard Visualization Research and Teaching Laboratory for co-creating the tabletop visualization platform in~\cref{fig:teaserfigure}, specifically Rus Gant, Kachina Studer, and Michael Quan. The authors declare no conflict of interest.




\ifCLASSOPTIONcaptionsoff
  \newpage
\fi



%
\bibliographystyle{IEEEtran}
\bibliography{references}
\clearpage
\renewcommand\thefigure{\thesection.\arabic{figure}}
\setcounter{figure}{0}
\appendices
\section{Dataset}\label{sec:appendix_dataset}

\subsection{Flood imagery}
\subsubsection{xbd2xbd image translation data}\label{sec:appendix_data_xbd2xbd}
The xbd2xbd dataset contains further details:
\begin{itemize}
    \item The dataset contains imagery from seven flood events from multiple regions and years: 
    \begin{itemize}
        \item Hurricane Harvey (hv) in Texas, US, 2017
        \item Hurricane Florence in the Carolinas, US, in 2018 (fl)
        \item Hurricane Michael in Mexico Beach, US, in 2018 (mi)
        \item Hurricane Matthew in North Carolina, US, in 2016 (ma) 
        \item Spring flooding in the Midwest, US, in 2019 (mw)
        \item Sunda strait tsunami in Indonesia in 2018 (in)
        \item South Asian monsoon floods in Nepal in 2017 (ne)
    \end{itemize}
    \item The dataset contains imagery of hurricanes, a spring flood, a tsunami, and a monsoon.
    \item 30\% of all imagery display standing flood (${\sim}1370$). The imagery of Harvey and Florence display the most standing flood.
    \item The training set contains imagery from all events.
    \item The validation set is composed of 216 image-triplets: 108 of each hurricane Harvey and  Florence. The validation set excludes imagery from hurricane Michael or Matthew, because the majority of tiles does not display standing flood. The 108 images are picked at random from the total set of images of each event. 
    \item We do not have a held-out test set. Primarily, this is due to the limited data size. But we also did not perform extensive hyperparameter optimization of the pix2pixHD model on the validation set and thus mostly avoid the risk of overfitting on the validation set.
\end{itemize}
We did not use digital elevation maps (DEMs), because our model is intended as visualization layer of existing flood models. Our model is not intended to correct bias in the flood model forecasts. Thus, the information of low-resolution DEMs should already be contained in the flood model that produces the input flood masks.

More open-source, high-resolution, pre- and post-disaster images can be found in unprocessed format on DigitalGlobe's Open Data repository~\cite{Digitalglobe_2020}. Furthermore, NOAA Emergency Response Imagery publishes high-resolution post-flood aerial imagery at $0.25-1m/px$ resolution~\cite{noaa23imagery}.
\begin{figure}[t]
  \centering
  \subfloat{
    \includegraphics [trim=0 0 0 0, clip, width=1.0\columnwidth, angle = 0]{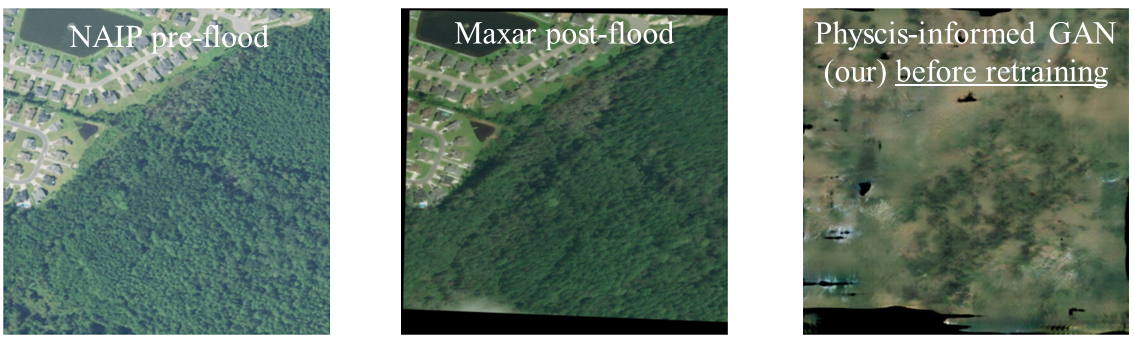}}
  \\
  \subfloat{
    \includegraphics [trim=0 0 0 0, clip, width=1.\columnwidth, angle = 0]{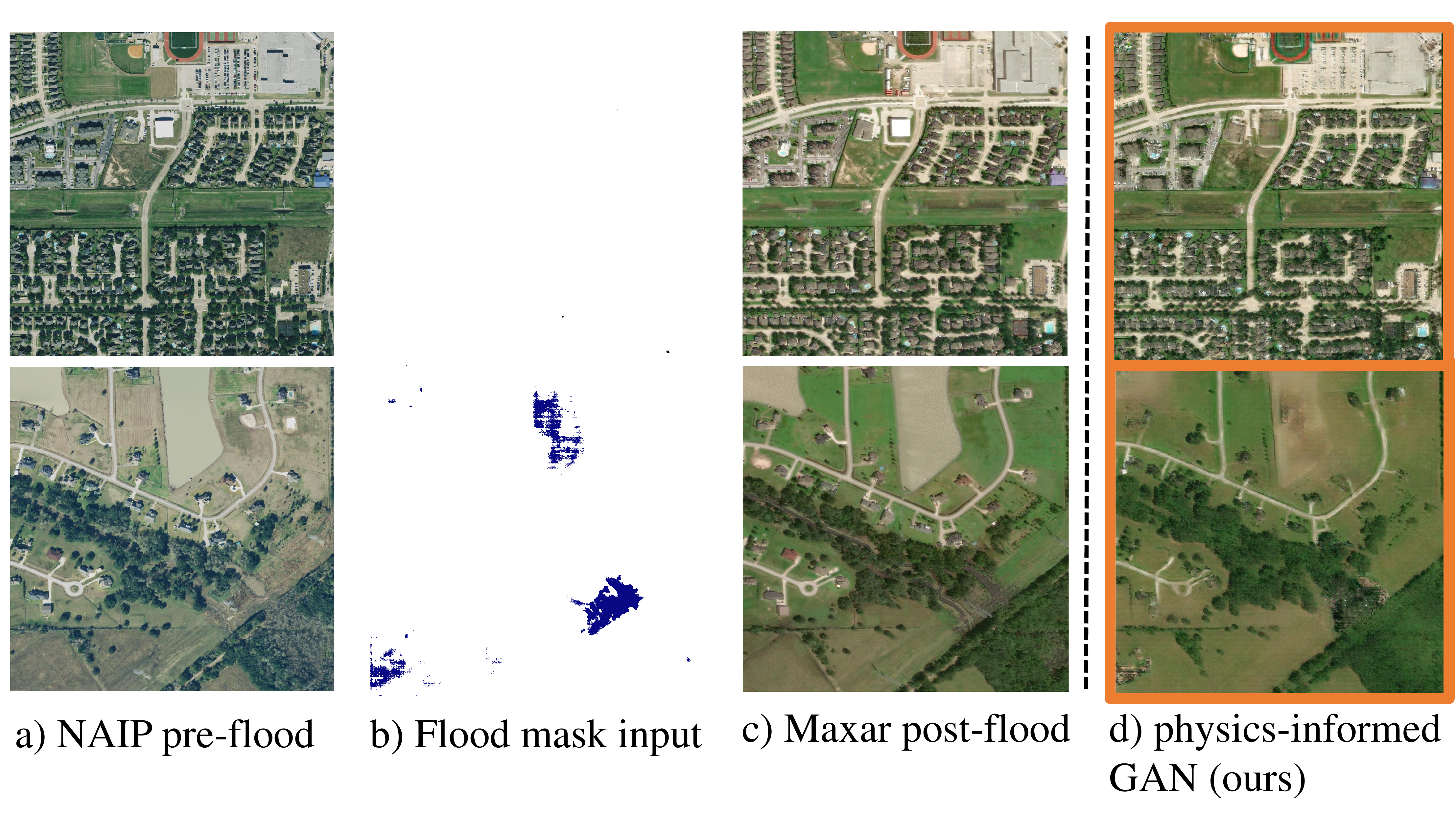}}
\caption[Generalization across remote sensing instruments]{\textbf{Generalization across remote sensing instruments.} We tested if a model can be trained on data from one remote sensing instrument and generalize to another. To do so, we trained a model on xbd2xbd and during testing changed the inputs from Maxar to NAIP imagery using naip2xbd. This experiment generated unintelligible imagery (top-row, right). This is likely due to the changes in resolution, color calibration, atmospheric noise, and more. To overcome this issue and train a model that generalizes across remote sensing instruments, we have compiled the naip2hou dataset of $5602$ image-triplets (a, c in two bottom rows). A model trained and tested on this dataset generates imagery in d) that is not production ready, but suggests that a retrained model can learn to translate images across different remote sensing instruments.} \label{fig:houston_west}
\end{figure}
\subsubsection{xbd-seg flood segmentation data}\label{sec:appendix_data_xbd_seg}
We hand selected 109 post-flood images from all events in the xbd2xbd training set to approximately maintain the distribution in event and percentage of flood extent. We labeled the images with floods that were contained by a convex hull by drawing polygon bounding boxes. We labeled the other imagery that contained objects inside the flood (houses, trees) with the help of the photoshop ``select areas of similar color``-tool. We used one annotator per image. The masks are binary thresholded and contain 0 for flood or water bodies, 1 for everything else, and 1 for no-value pixels. 

The data was randomly split into 5 equally sized partitions of $\approx$ 22 image-pairs to perform a 5-fold cross-validation. 

\subsubsection{Physics-based flood masks in xbdfathom}\label{sec:appendix_data_xbdfathom}

We retrieve a physics-based flood mask from the hurricane Harvey flood hazard simulations of the 30m-resolution US variant of the Fathom large-scale hydraulic modeling framework; downloadable at~\cite{wehner21fathomdata}. 
The hindcasts for hurricane Harvey are described in~\cite{wing19fathomharvey}, the US variant in~\cite{wing17fathomus}, and the original global hydraulic model framework in~\cite{sampson15fathomglobal}. The model simulates waterflow across a 2D land surface using the Venant shallow water equations and accounts for fluvial (riverine), pluvial (rainfall), and surge (coastal) flood hazards through appropriate boundary conditions.

The downloaded flood mask is a single-channel large tif for the Greater Houston area and contains the estimated flood height per ${\sim}30\unit{m/px}$ pixel. We binarize the mask by thresholding values above $20\unit{cm}$ following the protocol in~\cite{wehner21fathomharveyattribution} which mentions that residence damages usually start to occur at this height~\cite{uk19floodheight}. We cut the mask into tiles that match the xbd2xbd hurricane Harvey validation set pre-flood imagery by reprojecting the flood mask onto same coordinate system, resampling the mask to ${\sim}0.5\unit{m/px}$ via nearest neighbor interpolation, and then cropping out matching tiles. 

\subsubsection{Generalization data naip2xbd, naip2hou, and hou-seg}\label{sec:appendix_data_generalization}

We created the datasets naip2xbd, naip2hou, and hou-seg to test if our model can visualize flooding with inputs from the NAIP aerial imagery and targets from Maxar satellite imagery. 

The naip2xbd imagery was compiled by sourcing one NAIP imagery tile for every post-flood tile in the xbd2xbd dataset that belongs to the hv, fl, mi, or mw events. To do so, we extracted the geocoordinates for every flood tile in the xbd xview2 dataset. The xview2 dataset contained erroneous geoinformation for a few tiles, which we corrected by plotting all tiles. Using the extracted bounding boxes, we downloaded the NAIP imagery tiles using Google Earth Engine's javascript API. We upsampled the NAIP tiles from $1$m/px to $0.5$m/px. Further, there was distortion and an offset of 1-20m between the NAIP inputs and Maxar targets. To overcome this variance partially, we orthorectified the Maxar targets using SWIFT features, such that, both images are better aligned. In total, the dataset contains 6.1GB and 2063 HD image-triplets (1260 train, 394 test, and 409 held-out) using a random split.

The naip2hou and hou-seg contain post-flood images from a single swath of a Maxar satellite over West Houston after hurricane Katrina. The swath was sourced from Maxar's open data repository. We hand-selected areas that contained significant flooding, tiled the imagery of those areas, and hand-labelled a subset. We hand-labelled imagery with the help of the "select regions of similar color" tool in Adobe Photoshop. After tiling the dataset, we download teh associated NAIP imagery using Google Earth Engine. The hou-seg dataset contains 260 HD image-pairs and naip2hou contains 16GB of 5602 image-triplets (4481 train, 840 test, and 281 hold-out). 

\subsection{Pre- and post-reforestation imagery}\label{sec:appendix_data_forest}

\begin{figure}[t]
  \centering
  \includegraphics [trim=0 0 0 0, clip, width=1.\columnwidth, angle = 0]{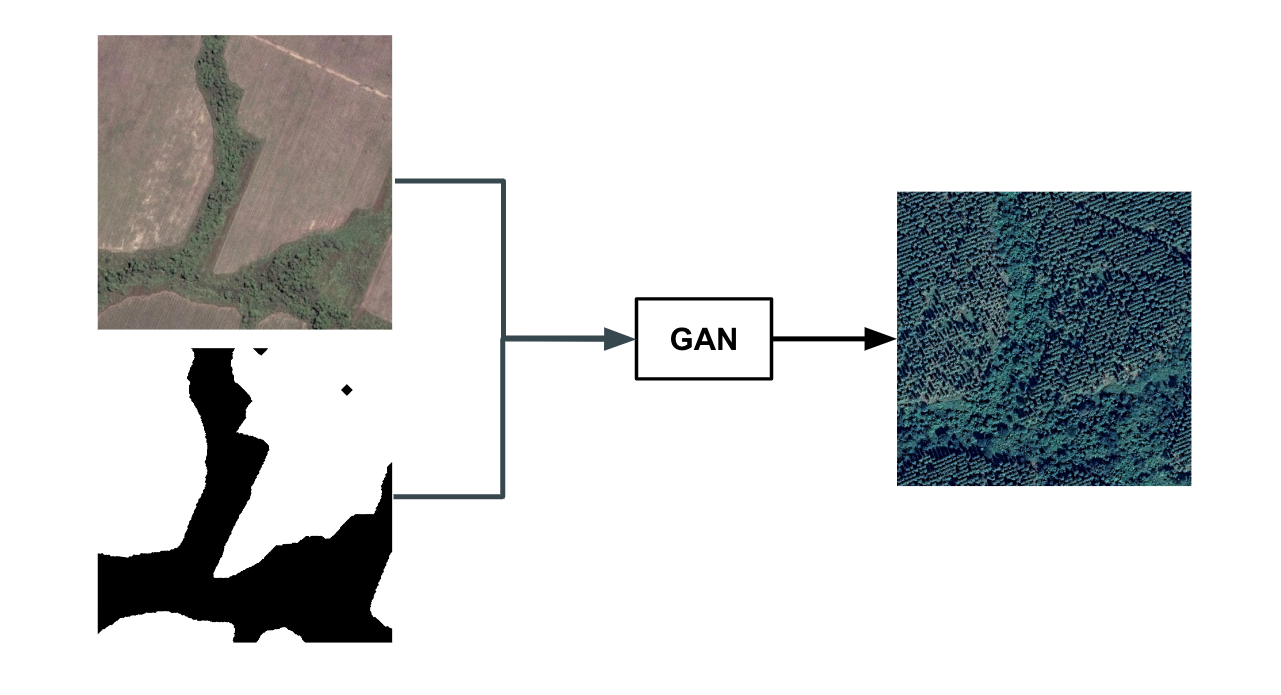}
\caption[Model architecture for visualizing reforestation]{\textbf{Model Architecture for Reforestation.} Our model leverages the semantic image synthesis model, Pix2pixHD, and combines a pre-reforestation satellite image with a reforestation map to generate post-reforestation satellite imagery.} \label{fig:model_architecture_forest}
\end{figure}

We compiled image triplets for visualizing reforestation, as seen in~\cref{fig:model_architecture_forest}. To select the areas, we went through all of VERA registered ARR (Afforestation, Reforestation, and Revegetation) carbon projects until 2022, and downloaded the shapefiles (.kmz or .kml) when they were available. We used Google Earth Pro to first confirm that cloud-free high-resolution ($<$3m/px) imagery for the years before project start was available over the shapefile regions. Then we visually verified that reforestation actually occured. These conditions were considered satisfied if we had imagery of bare land where we could see trees planted and grown over the years. In that case, we considered “pre-reforestation” as the earliest available imagery before we could see that trees were planted. For instance, if on the high-resolution imagery, we saw that trees were planted in 2010, then we would go back in time to the previously available imagery which could be 2009 or 2005, depending on the regions. For the “post-reforestation” imagery, we selected the most recent imagery available without clouds, so ideally it would be 2022. Our dataset timestamps go from 2005 to 2022 with some in-between years when those years were not available or not good quality (too many clouds, overlapping rasters from different years, etc).

\begin{figure}[t]
  \centering
  \includegraphics [trim=0 0 0 0, clip, width=1.\columnwidth, angle = 0]{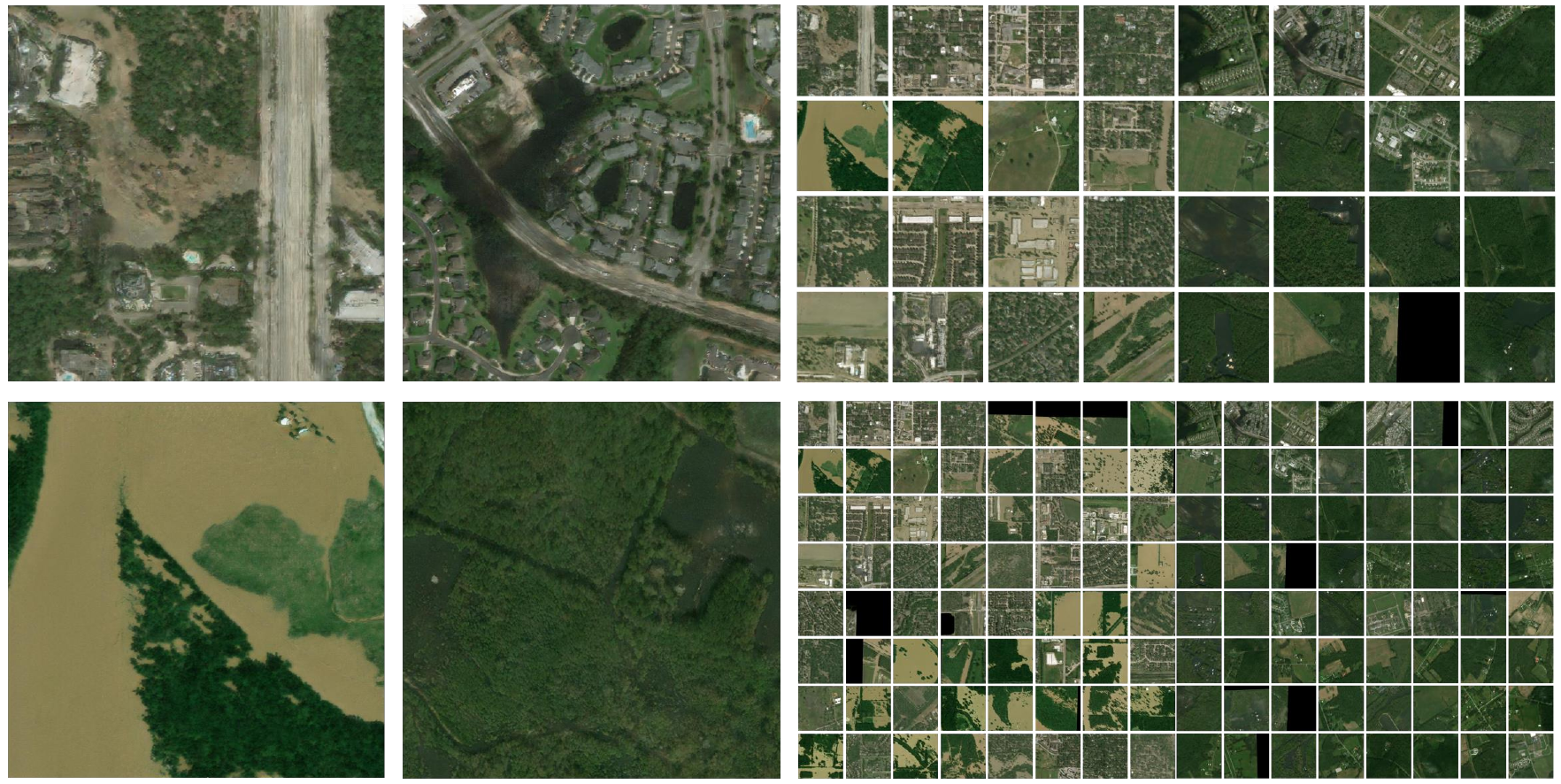}
  \vspace{.02in}
\caption[A collection of randomly generated flood images]{\textbf{Generated post-flooding imagery of $64$ randomly chosen tiles} using high-res. flood masks from hurricanes Harvey and Florence validation set.} \label{fig:grid_of_gen_ims} 
\end{figure}

Ultimately, the selected high-resolution images of before and after reforestation were exported from Google Earth Pro (Map data: Google, Maxar Technologies, CNS/Airbus) with a resolution of 4800x4800 and an eye altitude of 1500 meters. Our filenames follow the convention $<$grid reference time$> $\_eye\_alt\_ $<$eye\_altitude$>$m\_ $<$year\_aquisition$>$. For example, 
15QVV3535563291\_eye\_alt\_1500m\_2019. The image of the pre-reforestation with the visual of the shapefile layer was also exported to be used as the binary reforested area mask. The image with the visual of the shapefile was converted to a binary mask using an image processing method that removed all non-white pixels. 

The 1024x1024 tiles were generated from the 4800x4800 exported images. The binary reforestation area masks were stacked on top of the RGB  images. Only tiles that had pixels belonging to a reforested area were saved. In total, the forest dataset contains 5.5GB of 1272 image triplets (1026 train, 246 test) at a random 80/20 split across all collected regions and the forest-gtm datasets contains 460MB or 107 held-out image triplets from a previously unseen area in Guatemala.

\subsection{Pre- and post-melt Arctic sea ice imagery}\label{sec:appendix_data_arctic}
\begin{figure}[t]
  \centering
  \subfloat[Data Samples]{
    \includegraphics [trim=0 0 0 0, clip, width=1.\columnwidth, angle = 0]{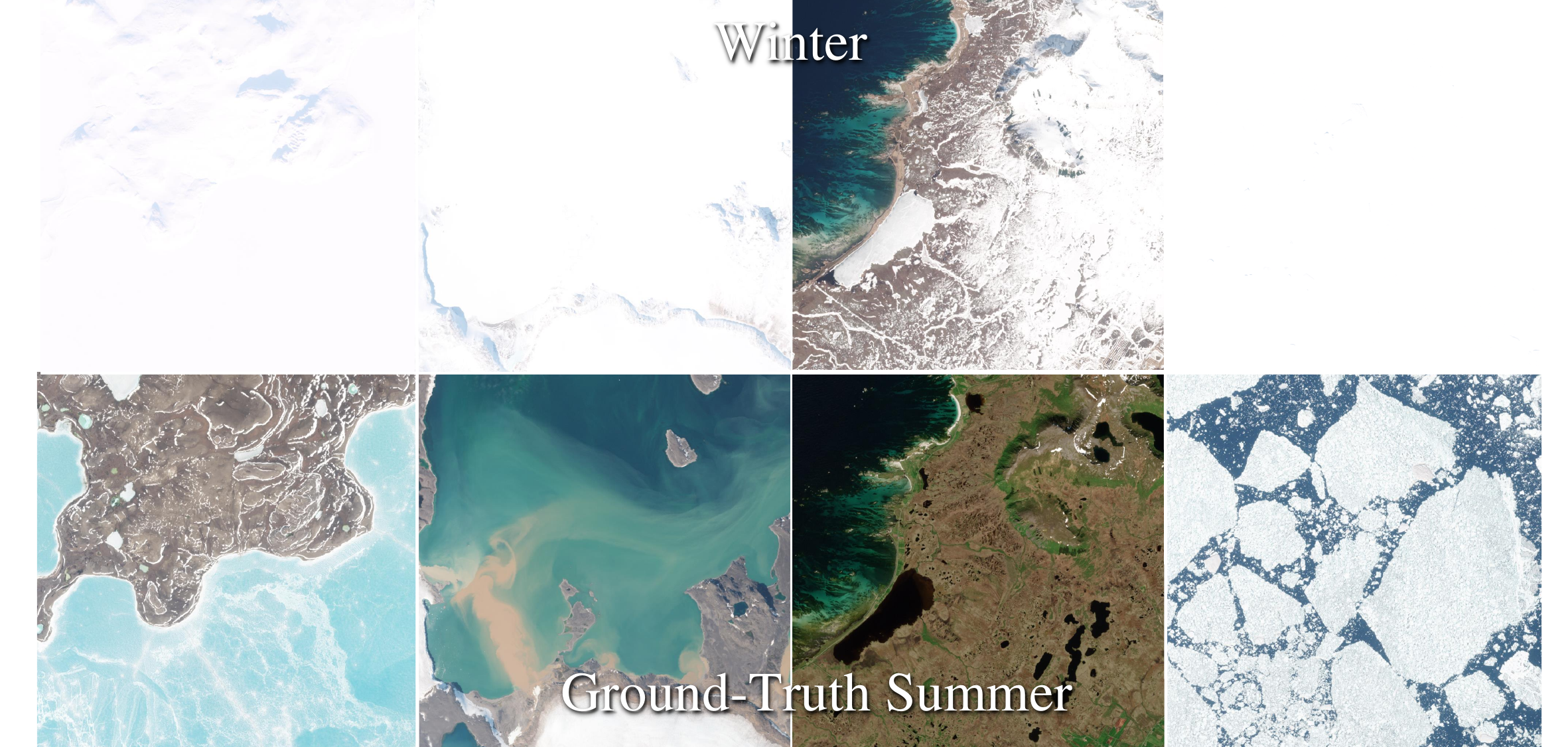}
    }
\caption[Data samples of melting Arctic sea ice]{We allow the extension of \eie/ to visualize melting Arctic sea ice by publishing an according dataset.} \label{fig:arctic_model} 
\end{figure}

\begin{figure}[t]
  \centering
  \subfloat[Study Area]{
    \includegraphics [trim=0 0 0 0, clip, width=0.95\columnwidth, angle = 0]{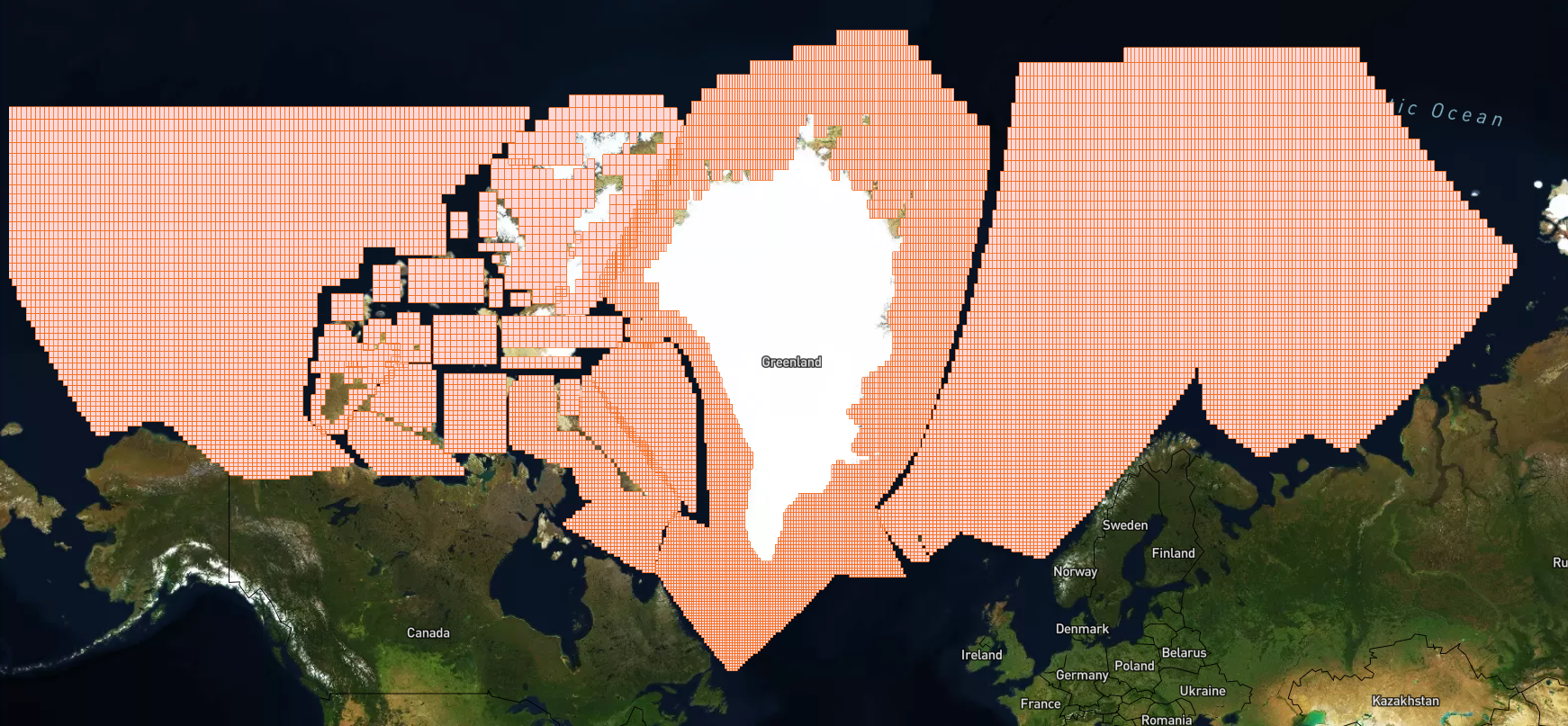}
    \vspace{0.05in}
    \label{fig:arctic_study_area}
    }
  \\
  \subfloat[Final data distribution]{
    \includegraphics [trim=0 0 0 0, clip, width=0.85\columnwidth, angle = 0]{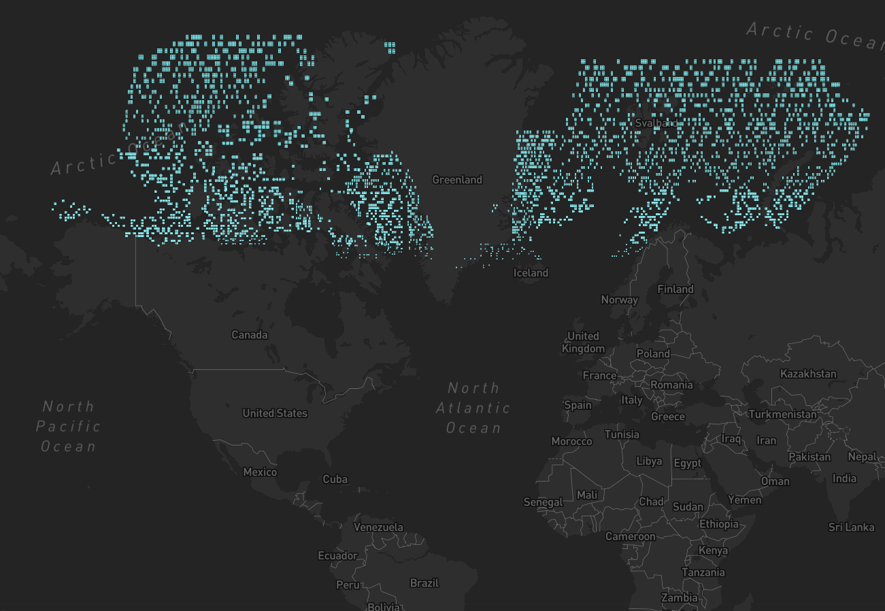}
    \vspace{0.05in}
    \label{fig:arctic_data_distribution}
    }
\caption[Data distribution of Arctic data]{We compiled a dataset of $\sim 20k$ image-pairs of the Arctic to visualize melting Arctic sea ice.} \label{fig:arctic_dataset} 
\end{figure}

To visualize melting Arctic sea ice we created 19445 image-triplets of pre-melt image, post-melt image, and post-melt segmentation mask, as displayed in~\cref{fig:arctic_model}. Because the retreat of Arctic sea ice occurs over decades, we used Winter and Summer imagery as pre- and post-melt, respectively. We found $\sim20k$ matching pre- and post-melt images by finding matching pairs across 27172 Winter images in 1st Oct. 2019 - 1st May 2020 and 32433 Summer images in 1st Jun. - 31st Aug. 2020 within the study area in~\cref{fig:arctic_study_area}. We downloaded cloud-free Sentinel-2 MSI Level 1-C visual imagery at 10m/px resolution in tiles of 1024x1024px that matched the criteria. 
Our ice segmentation model creates binary ice segmentation masks (1=ice) by thresholding grayscaled images into white and non-white (intensity $<255$) areas, \texttt{x = (1 if x==1 else 0)}. We created the post-melt segmentation masks by applying the segmentation model on the post-melt imagery. After creating segmentation masks, we rejected all image pairs that only contain ocean or land(mask=0).~\Cref{fig:arctic_data_distribution} shows the final locations of all image pairs. 

\begin{figure}[t]
  \centering
  \includegraphics [trim=0 0 0 0, clip, width=1.\columnwidth, angle = 0]{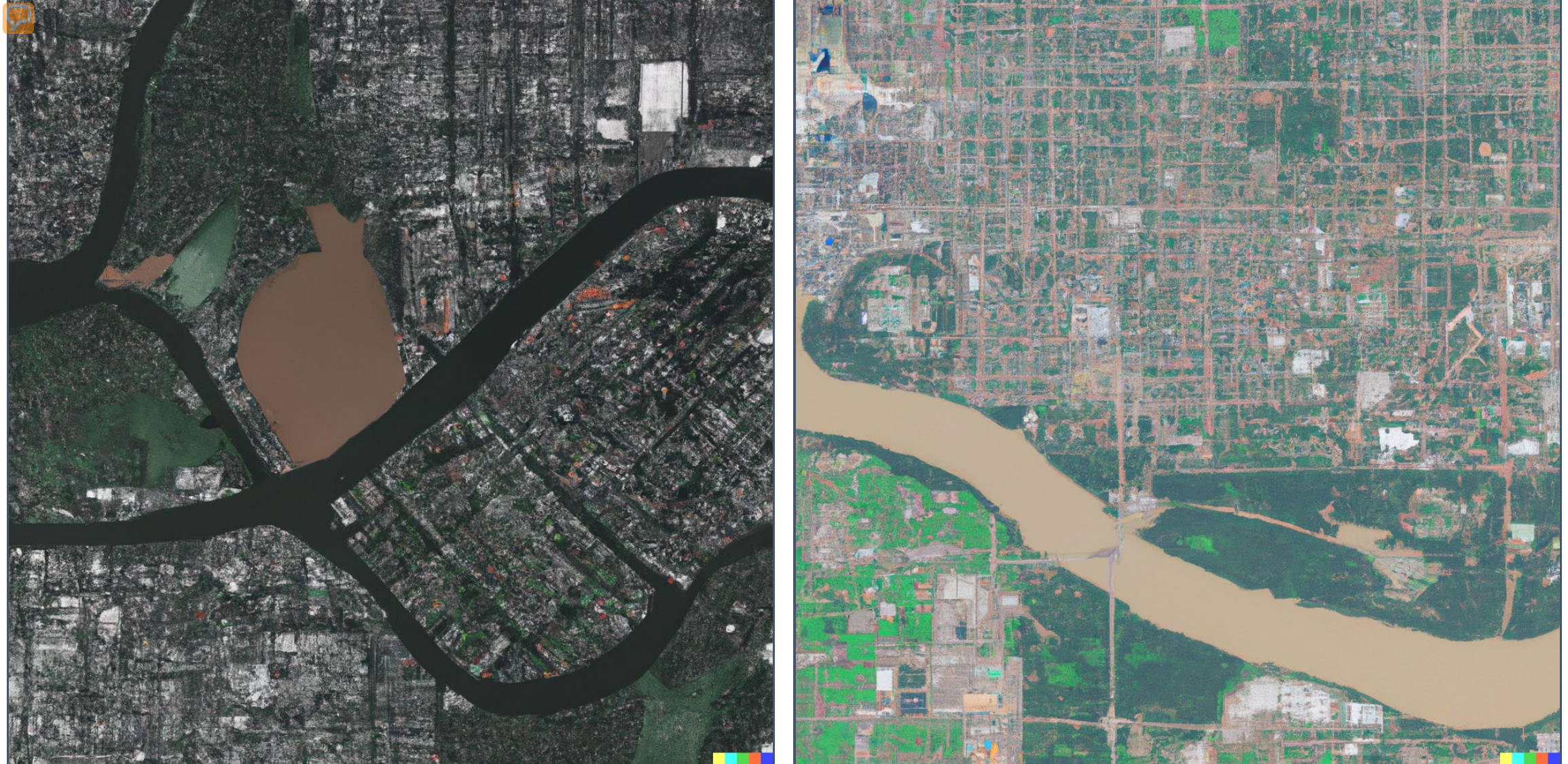}
  \vspace{.02in}
\caption[Dalle2 results on flood visualizations]{\textbf{Dalle-2.} We prompted the diffusion-based model, DALL·E 2, to generate ´A realistic and physically-consistent satellite image of Houston, TX being flooded.´ The visualized urban scenery is noisy and lacks the detail of buildings (zoom in). The blur is likely due to the complexity of generating the high-frequency and structured information that characterizes satellite images. This further indicates the continued challenge that generating satellite imagery, instead of drawings or first person images, is posing to deep generative vision models.} \label{fig:dalle2} 
\end{figure}

\section{Flood segmentation model}\label{sec:appendix_flood_seg}
We evaluate the physical-consistency of the generated flood imagery by measuring intersection over union (IoU). Specifically, we measure the IoU of a flood mask that is derived from the generated image via a flood segmentation model with the ground-truth flood mask that was used as input. Here, we provide additional methodology and results for the flood segmentation model.   

We trained two independent flood segmentation models -- one for the main xbd2xbd experiments in~\cref{sec:eie_results_main} using the 109 labelled images in xbd-seg and another on the 260 labeled images in the hou-seg dataset for~\cref{sec:gen_loc}. 
Our implementation is a pix2pix model~\cite{isola17pix2pix}, which differs from pix2pixHD~\cite{wang18pix2pixhd} and uses a vanilla UNet as generator. We trained it from scratch to minimize a modified loss function that minimizes $L1$-loss and IoU in addition to adversarial loss and finetuned the last layers on L1-loss. 

For both datasets, we used a UNet with 120 trainable layers to predict 1-channel segmentation masks from 3-channel images of size 1024x1024. The loss function is a weighted sum (1:1:5) of a vanilla GAN loss, L1 loss, and negative IoU wrt. the ground-truth segmentation mask. We trained the model for 80 epochs with batch size, 8, and learning rate, 0.0002, on 4 GPUs. For the xbd-seg dataset, we then fix the first 100 layers and fine-tune the last 20 layer network for 80 epochs using only L1 and IoU loss (0:1:3) and otherwise the same hyperparameters. 

We experimented with decreasing the number of epochs and removing the GAN loss to train a UNet on L1+IoU, which both individually decreased performance. Otherwise, we are using the default parameters in~\url{https://github.com/junyanz/pytorch-CycleGAN-and-pix2pix} master branch on 09/2020. For the hou-seg dataset, we re-trained the same network from scratch on the hou-seg data and did not fine-tune the network. 

The xbd-seg dataset is relatively small with 109 image-pairs. This data sounds extremely limited, but note that 109 images at HD resolution would equal 1744 image-pairs at 128x128 resolution. Nevertheless, to overcome the data limitation, we use a 5-fold cross validation as train-test split. I.e., we randomly split the dataset into 5 equally sized partitions of $\approx 22$ image-pairs and train the model with considered hyperparameters 5-times on 4 partitions while holding out 1 partition. After finding the model with the best hyperparameters, we train it on the full labeled dataset and use it to infer flood masks on all data in xbd2xbd. The model for hou-seg was trained using a single random 80-20 train-test split and otherwise same hyperparameters as for xbd-seg. 

The model on the xbd-seg dataset has IoU 0.343 and on the hou-seg has IoU 0.23. Many images in our dataset display very little flooding and, hence, achieving high IoU scores is very difficult. The IoU of 0.343 matches expections of the pix2pix model~\cite{isola17pix2pix}. For a better understanding that the IoU of 0.23 on hou-segshould be sufficient, we are plotting 20 randomly selected samples from the hou-seg test set in~\cref{fig:results_flood_seg}.

Future work, could likely improve the IoU by choosing a network backbone that is pretrained on visual satellite imagery, more complex segmentation models~\cite{paperswithcode23cityscapes}, or increasing the data size.

The segmentation model could be improved, e.g., by considering more novel segmentation architectures, such as, PSPNet~\cite{zhao17pspnet}, PAN~\cite{li18pan}, or DeepLabv3+\cite{chen18deeplabv3plus}. Due to the limited amount of data, we expect semi-supervised learning approaches to impact the performance more than architectural choices. Thus, we experimented with networks that were pretrained on ImageNet, but they classified "background" for every image, which led us to believe that mid-to-late-stage layers have little to no signal and only early layers would be useful. Future work, could use networks that are pretrained via contrastive or reconstruction loss on other remote sensing tasks and datasets, similar to~\cite{klemmer22thesis,lacoste21geobench}. 

\section{Experiments}\label{sec:appendix_experiments}

\subsection{Data Augmentation.}\label{sec:appendix_augmentations}
To visualize floods, we applied standard data augmentation, here rotation, random cropping, hue, and contrast variation, and state-of-the art augmentation - here elastic transformations~\cite{simard2003best}. Furthermore, spectral normalization~\cite{miyato2018spectral} was used to stabilize the training of the discriminator. A relativistic loss function has been implemented to stabilize adversarial training. We also experimented with training pix2pixHD on LPIPS loss. Quantitative evaluation of these experiments, however, showed that they did not have significant impact on the performance and, ultimately, the results in the paper have been generated by the pytorch implementation of pix2pixHD~\cite{wang18pix2pixhd} extended to $4$-channel inputs. 

To visualize reforestation, we used downscale (to 0.8 scale), h- and v-flip, and colorjitter (brightness=0.4, contrast=0.2, saturation=0, hue=0) augmentations with $p=0.67$ from the albumentations library~\cite{buslaev18albumentations}. The model hyperparameters are chosen to equal the pytorch implementation in~\cite{wang18pix2pixhd}.

\subsection{Pre-training LPIPS on satellite imagery.} The standard LPIPS did not clearly distinguish in between the handcrafted baseline and the physics-informed GAN, contrasting the opinion of a human evaluator. This is most likely because LPIPS currently leverages a neural network that was trained on object classification from ImageNet. The neural network might not be capable to extract meaningful high-level features to compare the similarity of satellite images. In preliminary tests, we ran inference using an ImageNet-pretrained network and saw that it classified all satellite imagery as background image, indicating that the network did not learn features to distinguish satellite images from each other. Future work, will use LPIPS with a network trained to have satellite imagery specific features, e.g., Tile2Vec or a land-use segmentation~\cite{Robinson_2019} model. 

\section{Results}\label{sec:appendix_results}

\begin{figure*}[t]
  \centering
  \begin{subfigure}{1.\textwidth}
      \centering
      \includegraphics [trim=0 0 0 0, clip, height=\dimexpr
  \textheight-5\baselineskip-\parskip-.2em-
  \abovecaptionskip-\belowcaptionskip\relax, angle = 0]{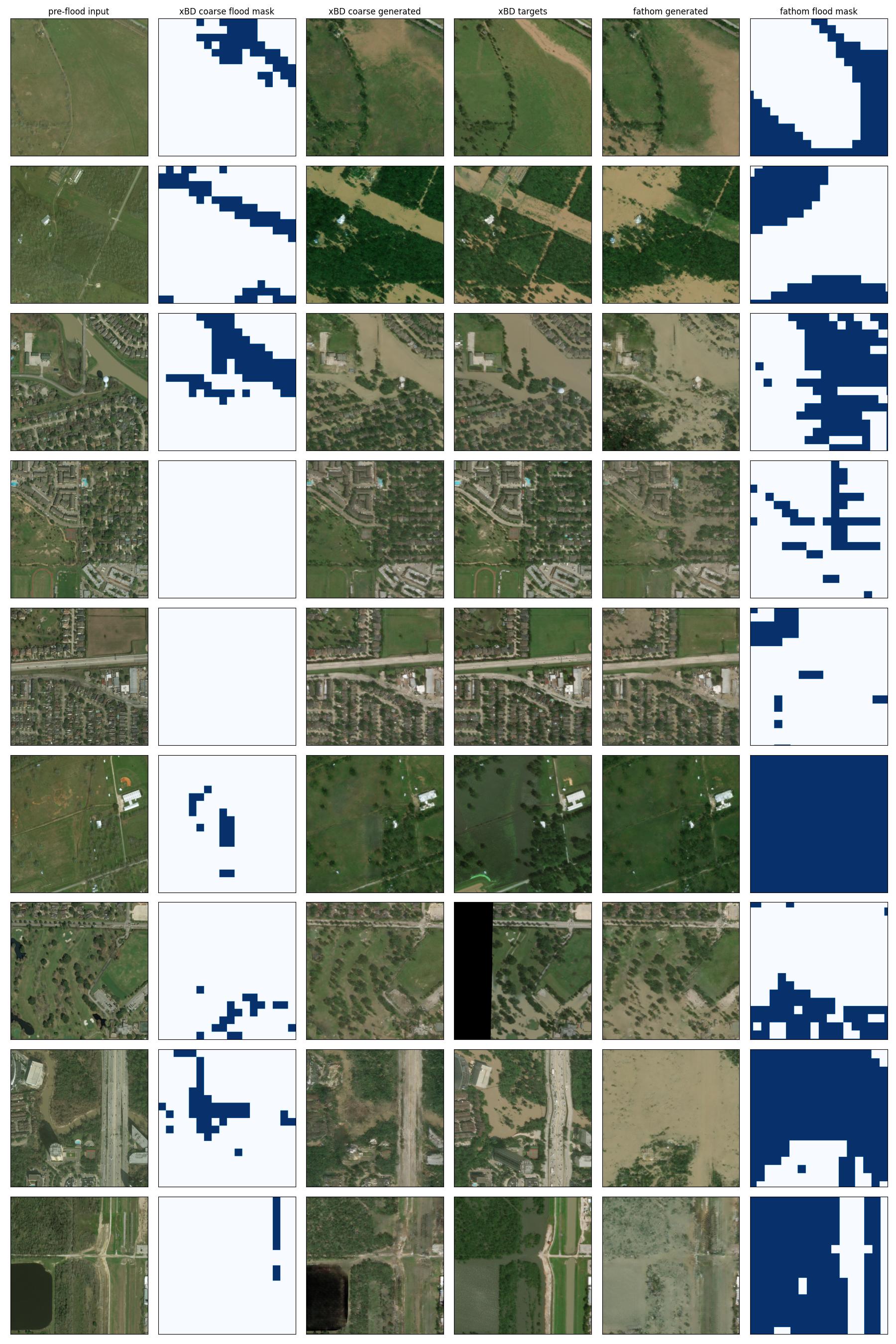}
  \end{subfigure}
\caption[xbdfathom generated images]{The plot shows inputs, targets, and predictions from the xbd2xbd hurricane Harvey validation set (1st-4th col.) and the xbdfathom dataset (5th-6th. col.). The selected imagery in the first two rows displays how differing flood mask inputs (2nd and 6th col.) affect the generated visualization (3rd and 5th col.) if the pre-flood input image (1st col.) and model are held fixed. The next five rows are randomly selected and the last two rows display selected failure cases.} 
\label{fig:xbdfathom_generated_images} 
\end{figure*}

~\Cref{fig:xbdfathom_generated_images} shows inputs, targets, and predictions from the xbd2xbd and xbdfathom dataset. The two top rows show selected success cases. The model accurately in-paints a brown-colored flood (3rd and 5th col.) in the areas of the xbd2xbd (2nd col.) and xbdfathom (6th col.) flood masks. Many high-resolution features, such as the location of the houses, trees, or roads seem to be maintained across pre-flood (1st col.) and generated image (3rd and 5th col.); matching the expectations from the xbd2xbd target imagery (4th col.). In many locations within the images, the model seems to correctly visualize a flood height that is lower than house roofs or forest canopy (except for, e.g., top-left area in 5th col., 2nd row).

The next five rows show imagery that is randomly selected from the xbd2xbd test dataset. Overall, there does not seem to be a substantial difference in quality in between the visualizations of the xbd2xbd observation-derived (2nd col.) or xbdfathom physics-based (6th col.) flood masks. Similar to the success cases, the model seems to maintain the location of many houses in residential areas, but also overpaints structure occasionally with a flood-brown (e.g., 5th col., 3rd row, bottom-left area). The model seems to ignore masks that contain floods in every pixel (e.g., 5th col., 6th row), which possibly suggests that learned weights are only activated by spatial gradients in the flood mask.

The two bottom rows show failure cases: The penultimate row indicates that the model fails to capture high-rise buildings, cars, or bridges, which are features that occur rarely in the training dataset. The last row shows that the low-res. flood masks in xbd2xbd can contain partially inaccurate labels, in this case missing the standing flood in the forested area (comparing 2nd to 4th col.). The last row also shows a failure case where the model visualizes an unnatural flood color with many distortions (5th col.).

\begin{figure*}[ht]
  \centering
  \includegraphics [trim=0 0 0 0, clip, height=0.9\textheight, angle = 0]{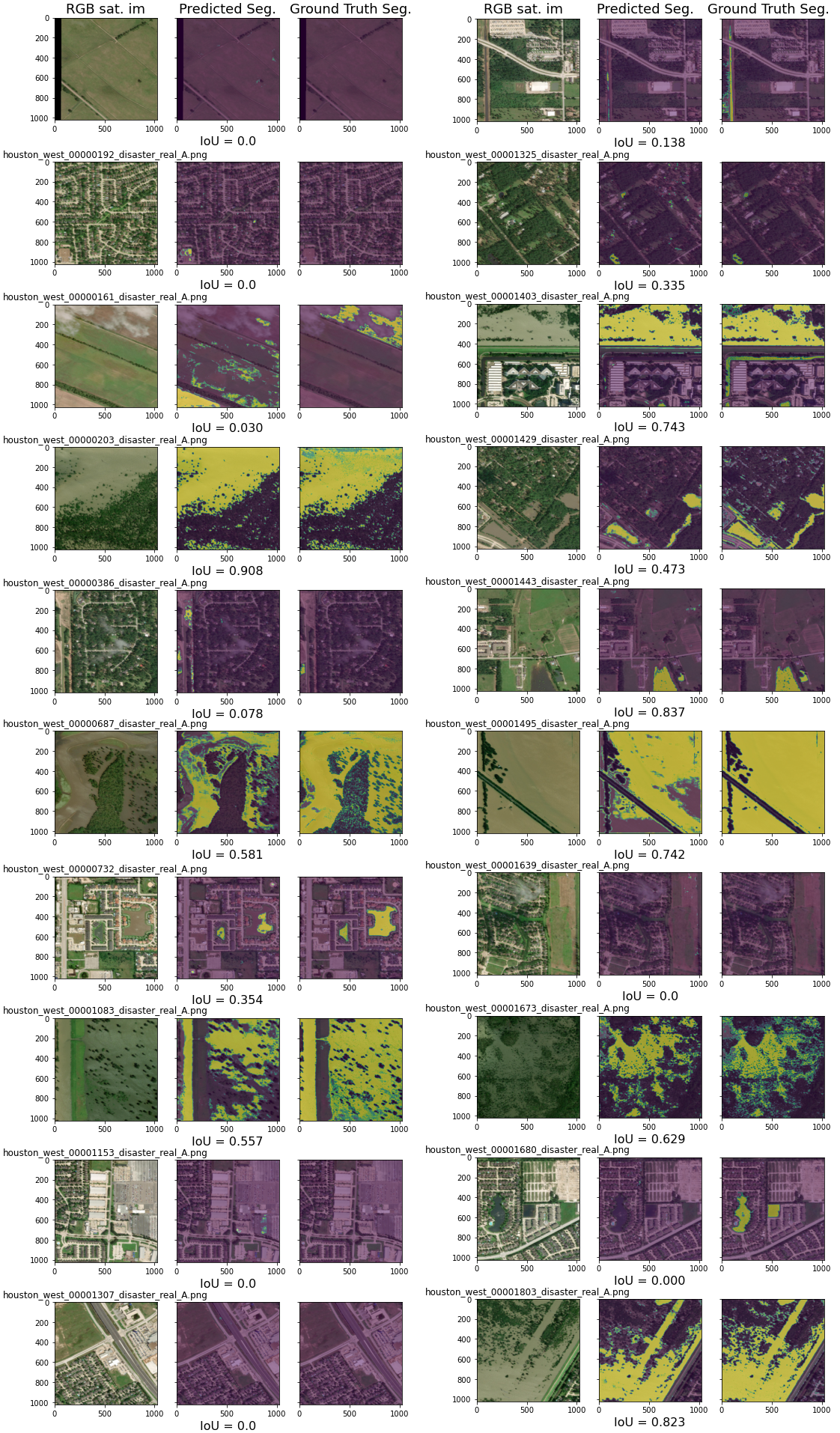}
      \vspace{.02in}
\caption[Flood segmentation model]{Generated segmentation masks for 20 randomly  chosen tiles of naip2hou dataset (zoom in). The average IoU across the dataset is 0.23.} \label{fig:results_flood_seg} 
\end{figure*}

We attach an additional 20 randomly chosen prediction from the naip2hou dataset in~\cref{fig:results_flood_seg} to illustrate the performance of the naip2hou flood segmentation model. Each image triplet shows the input image (left), prediction (middle), and ground-truth (right). The mean IoU of our model over the validation dataset is 0.23, which is below average performance of a segmentation model. But, note that the IoU for images with significant flooding is generally $>0.4$ and the IoU for images with near-zero water is $0.0$ (e.g., in the 2nd to bottom row). The images with near-zero flooding skew the reported average IoU towards $0$. 

\begin{figure}[H]
  \centering
  \begin{subfigure}{1.\columnwidth}
      \centering
      \includegraphics [trim=0 0 0 0, clip, width=0.98\textwidth, angle = 0]{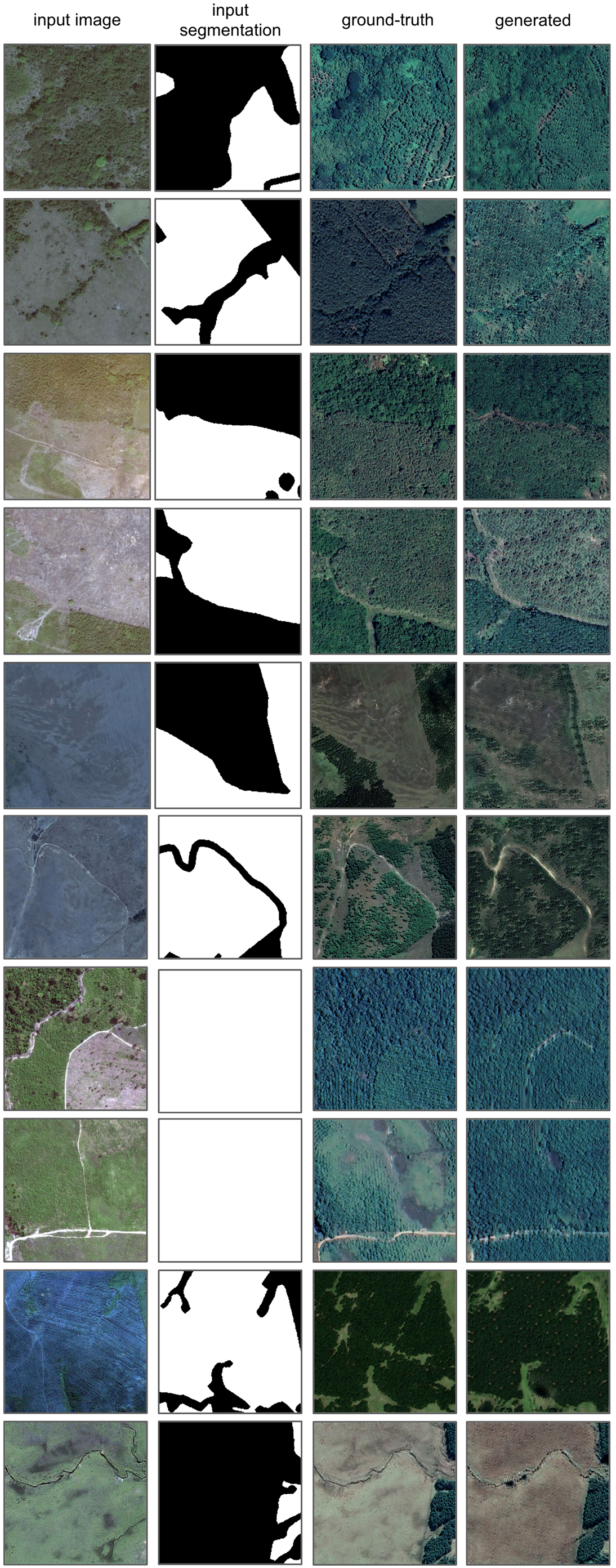}
  \end{subfigure}
  \caption[Forest random results]{Generated visualizations for 10 randomly chosen tiles of the \textbf{forest} validation dataset. The white areas in the input segmentation indicates areas in the ground-truth image that have been reforested after the input image was taken.} 
    \label{fig:results_forest_random_10} 
\end{figure}

We also plot 10 randomly chosen images from the forest validation dataset in~\cref{fig:results_forest_random_10}.

\end{document}